%% file: main.tex
\documentclass[10pt,twocolumn,letterpaper]{article}

\usepackage[pagenumbers]{cvpr} 

\newcommand{\methodabbr}{ProtoDepth\xspace}

\input{preamble}

\definecolor{cvprblue}{rgb}{0.21,0.49,0.74}
\usepackage[pagebackref,breaklinks,colorlinks,allcolors=cvprblue]{hyperref}

\def\paperID{3351} 
\def\confName{CVPR}
\def\confYear{2025}

\title{ProtoDepth: Unsupervised Continual Depth Completion with Prototypes}

\author{Patrick Rim\;\;\;Hyoungseob Park\;\;\;S. Gangopadhyay\;\;\;Ziyao Zeng\;\;\;Younjoon Chung\;\;\;Alex Wong \vspace{3mm} \\
Yale Vision Lab 
}

\begin{document}
\maketitle

\begin{abstract}
We present ProtoDepth, a novel prototype-based approach for continual learning of unsupervised depth completion, the multimodal 3D reconstruction task of predicting dense depth maps from RGB images and sparse point clouds.
The unsupervised learning paradigm is well-suited for continual learning, as ground truth is not needed.
However, when training on new non-stationary distributions, depth completion models will catastrophically forget previously learned information.
We address forgetting by learning prototype sets that adapt the latent features of a frozen pretrained model to new domains.
Since the original weights are not modified, \methodabbr does not forget when test-time domain identity is known.
To extend \methodabbr to the challenging setting where the test-time domain identity is withheld, we propose to learn domain descriptors that enable the model to select the appropriate prototype set for inference.
We evaluate \methodabbr on benchmark dataset sequences, where we reduce forgetting compared to baselines by 52.2\% for indoor and 53.2\% for outdoor to achieve the state of the art.
Project Page: \href{https://protodepth.github.io/}{\texttt{https://protodepth.github.io/}}.

\end{abstract}
\vspace{-1mm}

\section{Introduction}

Reconstructing three-dimensional (3D) environments facilitates spatial applications such as autonomous navigation, robotic manipulation, and AR/VR.
These applications typically utilize platforms equipped with heterogeneous sensors, such as cameras and sparse range sensors (e.g., LiDAR, ToF) which yield RGB images and point clouds.
To reconstruct 3D scenes from an ego-centric point of view in the form of depth maps, the point clouds are typically projected onto the image plane.
However, their projections tend to be sparse, which is unlike RGB images that comprise irradiance measured at each pixel.
These complementary modalities can be used to predict dense depth maps; this multimodal 3D reconstruction task is called depth completion. 

Depth completion models can be trained in a supervised (using ground truth) or unsupervised (using Structure-from-Motion) manner. As ground truth is prohibitively expensive to acquire, we subscribe to the unsupervised learning paradigm, which enables one to learn without human intervention. While this suggests the potential to continuously learn, existing models are trained and evaluated on single datasets under the assumption of a stationary data distribution. However, sequences of multiple datasets exhibit non-stationary distributions and are captured by sensors with varying calibrations. Hence, fitting to new data samples inevitably causes the model to ``catastrophically forget''~\cite{french1999catastrophic, mccloskey1989catastrophic, thrun1995learning, ratcliff1990connectionist} previously learned information, where the model performance degrades significantly on data from distributions that it had already observed.

\begin{figure}[t]
  \centering
  \includegraphics[width=\columnwidth]{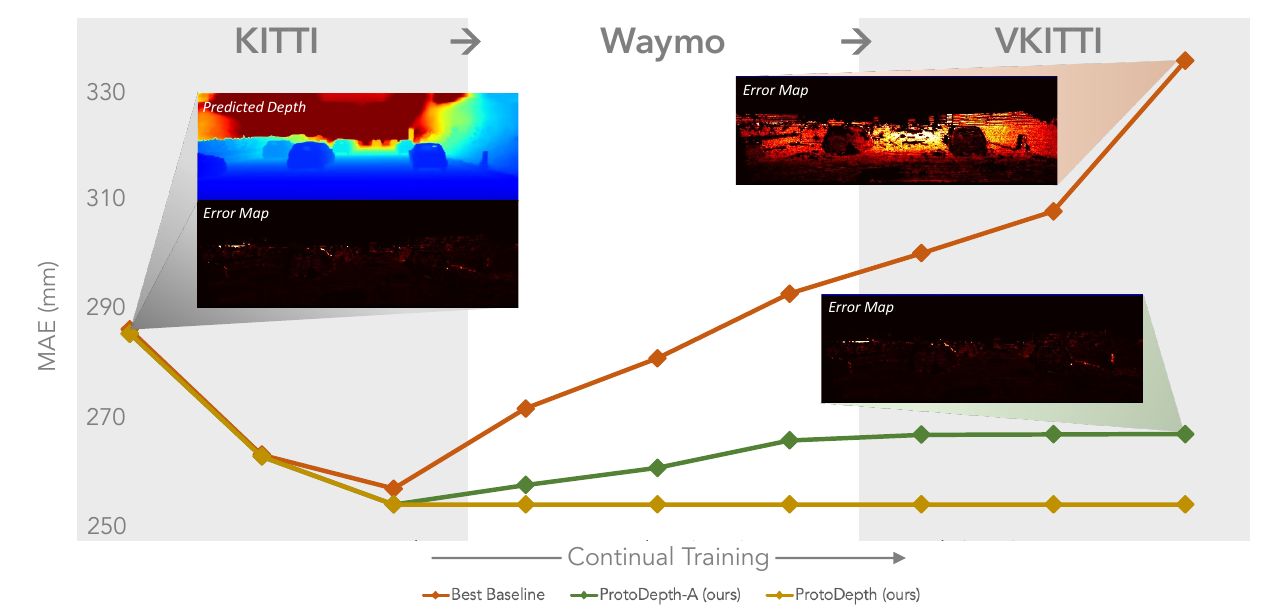}
  \vspace{-5mm}
  \caption{Results on \textbf{KITTI} validation set during continual training on outdoor dataset sequence (KITTI $\rightarrow$ Waymo $\rightarrow$ VKITTI).}
  \label{fig:teaser}
  \vspace{-2mm}
\end{figure}

To enable pretrained models to adapt to new environments or domains in an unsupervised manner, we consider continual learning, where training strategies aim to mitigate catastrophic forgetting of previously observed training distributions when learning from a continuous stream of non-stationary data. One school of thought is to constrain updates to parameters deemed ``important'' for previous domains based on a regularization term controlled by a weight chosen as a hyperparameter. Another is to incur memory costs by storing previously observed training data and incorporating them during learning of new domains, akin to rehearsals. The trade-off is in the sensitivity of optimization (and forgetting) to the choice of hyperparameter and in the size of storage and data privacy. Both schools perform updates to the full set of model parameters, which is computationally expensive. Unlike them, we opt to freeze the pretrained model to ensure zero forgetting and learn a small set of parameters, which we term ``prototypes,'' that modulate the latent features for each new domain encountered.  

We model the change in distribution as a domain-specific bias to be learned by global multiplicative and local additive prototypes that transform the latent features to fit the new distribution. In the same vein are learnable prompts or tokens~\cite{wang2022learning} used in continual learning for vision transformers, where the model itself is similarly frozen. Yet, there is no natural scale at which to discretize images, unlike discrete text tokens. In contrast, our proposed prototypes lift the requirements of tokenized inputs and can thus be applied to convolutional neural networks, which are prevalently used in unsupervised depth completion, as well as transformers.

To this end, we propose \methodabbr, a novel prototype-based method for unsupervised continual depth completion. Instead of prompts in input space, we deploy lightweight prototypes to a frozen pretrained model to encode prototypical information of each domain. These prototypes model global and local biases, where global prototypes learn a transformation from the latent pretrained data distribution to that of the new domain, and local prototypes capture fine-grained features that can be selectively queried depending on the input. Naturally, when the test-time domain identity is known, i.e., domain-incremental, \methodabbr exhibits no forgetting and learns the new data distribution with high fidelity. We further encode each domain as a descriptor to enable inference when test-time domain identity is withheld, i.e., domain-agnostic, where the prototype set corresponding to the highest affinity domain descriptor for a given sample is chosen.

\textbf{Our contributions}: We propose (1) a novel prototype-based paradigm for unsupervised continual depth completion that incurs no forgetting in the domain-incremental setting, and (2) a prototype set selection mechanism that extends the prototype paradigm to domain-agnostic settings with minimal forgetting. This is facilitated by (3) a novel training objective that learns descriptors for each domain, which can be used to determine the prototype set suitable for inference without knowledge of domain identity. (4) Our method, \methodabbr, reduces forgetting over baselines by over 50\% across six datasets; to the best of our knowledge, this is the first unsupervised continual depth completion method.

\section{Related Work}

\textbf{Continual learning} is the process of incrementally adapting the weights of a parameterized model to perform new tasks involving non-stationary distributions, while preserving information learned from previous tasks.

\textit{Regularization-based} methods aim to mitigate forgetting by restricting the plasticity of model parameters that are important for previously learned tasks. \cite{kirkpatrick2017overcoming, zenke2017continual, aljundi2018memory, chaudhry2018riemannian} introduce regularization terms to the loss function to identify and penalize changes to important weights while new tasks are learned.
\cite{rios2018closed, dhar2019learning, lee2019overcoming, zhai2019lifelong, douillard2020podnet, hou2019learning, li2017learning} attempt to preserve output behavior on previous tasks using knowledge distillation~\cite{hinton2015distilling, romero2014fitnets}. 
\cite{rannen2017encoder, nguyen2018variational, titsias2019functional, pan2020continual} regularize the space of learned functions to encourage similarity between outputs of task-specific heads.
However, while they perform well in simpler continual learning settings, regularization-based methods can struggle with more challenging tasks~\cite{mai2022online} and larger domain shifts between datasets~\cite{rebuffi2017icarl, wu2019large}.

\textit{Rehearsal-based} methods use a memory buffer to store a limited amount of data from previous tasks, allowing the model to periodically re-train on this data during continual learning. \cite{chaudhry2018riemannian, rolnick2019experience, hayes2019memory} introduce the strategy of retaining a subset of previous ``experiences'' (i.e., data) to ``replay'' (i.e., re-train on) while learning new tasks. 
\cite{chaudhry2019tiny, riemer2018learning, vitter1985random, lopez2017gradient, rebuffi2017icarl, kang2024continual} employ various sampling strategies, \cite{zhu2022self, liu2020generative, iscen2020memory} reduce memory by storing latent features rather than inputs, and generative replay methods~\cite{shin2017continual, ostapenko2019learning, ayub2021eec, wu2018memory, riemer2019scalable, rostami2019complementary, pfulb2021continual, kemker2018fearnet, gopalakrishnan2022knowledge} create synthetic rehearsal data from previous task domains.
Some recent works utilize techniques such as self-supervision~\cite{cha2021co2l, pham2021dualnet} and knowledge distillation~\cite{rebuffi2017icarl, wu2019large, chaudhry2021using, buzzega2020dark}.
\cite{chawla2024continual} introduces a rehearsal-based continual single image-based depth estimation method, the first such work for a 3D vision task. 
Rehearsal-based methods can reduce forgetting but are unsuitable when data storage is limited by memory or privacy constraints~\cite{shokri2015privacy}. Additionally, their performance degrades significantly as memory buffer size shrinks~\cite{cha2021co2l}.

\textit{Architecture-based} methods~\cite{mallya2018packnet, serra2018overcoming, wang2020learn, ke2020continual, rusu2016progressive, yoon2017lifelong, li2019learn, loo2020generalized, rao2019continual, zhao2022deep, kim2023achieving} allocate task-specific parameters or sub-networks, aiming to enable learning of new tasks while minimizing changes to parameters assigned to previous tasks.
Most of these methods require task identity to be known at test-time, which prevents their use in the more realistic domain-agnostic setting where new data is not restricted to a certain domain. Other methods~\cite{douillard2022dytox, wang2022continual} rely on a rehearsal buffer in addition to task-specific parameters. Furthermore, architecture-based methods often introduce a significant number of additional parameters for each task~\cite{wortsman2020supermasks, yan2021dynamically, pham2021dualnet}, which can even exceed the parameter count of the original model~\cite{wang2020learn, ke2020continual}.
In contrast, our method can perform inference without task identity, does not require a rehearsal buffer, and adds <5\% of the original model's parameters per task.

\textit{Prompt-based} methods introduce learnable prompts that encode task-specific information.
\cite{wang2022learning} learns a pool of tokens, from which a set is selected using a query mechanism and prepended to the input. 
\cite{wang2022dualprompt} refines this by using both task-specific and shared prompts. 
Subsequent approaches replace prompt selection with an attention mechanism~\cite{smith2023coda} or with intermediate embeddings~\cite{kim2024one}.
However, these prompt-based methods are designed for 2D classification tasks that use vision transformers (ViTs), borrowing the concept of prompting from the field of natural language processing (NLP). 
The idea of prepending prompts to tokenized inputs does not naturally extend to convolutional neural networks (CNNs), limiting their applicability to 3D vision tasks where CNNs are primarily used.
In contrast, our method learns prototypes, which serve as representative \emph{features}, offering a more intuitive mechanism for adding a lightweight selective bias than prepending abstract prompts in image space. Unlike prompt-based methods, our method is fully architecture-agnostic and can be applied to any model that has a latent space without modifying the underlying architecture.

\textbf{Prototypes} have predominantly been explored in the context of few-shot learning of 2D tasks~\cite{snell2017prototypical, liu2022dynamic, yang2020prototype, liu2020prototype, garcia2018few, gidaris2018dynamic} and representation learning~\cite{chen2019looks, caron2020unsupervised, li2020prototypical}. For continual learning, prototypes have been deployed for supervised classification tasks as discriminative class representations~\cite{wei2023online, asadi2023prototype, de2021continual, ho2023prototype, zhang2019variational, li2024steering}. In contrast, we leverage prototypes as a mechanism for encoding domain-specific information to facilitate continual learning for unsupervised reconstruction.

\textbf{Unsupervised depth completion} is a multimodal, 3D reconstruction problem of predicting a depth map from an image and its associated sparse point cloud. In lieu of training signals from ground-truth depth annotations, unsupervised depth completion methods~\cite{ma2019self, shivakumar2019dfusenet,wong2021learning,wong2021adaptive,wong2020unsupervised,yan2023desnet,yang2019dense,jeon2022struct,liu2022monitored,lao2024depth,wong2020targeted,ezhov2024all, singh2023depth, fei2019geo,wong2019bilateral}
minimize the reconstruction error of the sparse depth, the reprojection error of the image from temporally adjacent views, and a local smoothness regularizer.
Some works~\cite{yang2019dense, lopez2020project, upadhyay2023enhancing} utilize synthetic data to learn priors, while other works~\cite{wong2020unsupervised, wong2021learning} densify then refine the sparse point cloud by constructing a coarse piecewise scaffolding.
\cite{wong2021unsupervised} uses backprojection layers to map image features to the 3D scene using camera calibration. 
These methods are prone to catastrophic forgetting in continual learning settings since they must update their learned model weights to fit to new data, causing performance degradation on previously seen data domains.
In contrast, our method freezes the entire model and learns a lightweight selective bias for each dataset, allowing it to adapt effectively to new data domains while mitigating forgetting. 

UnCLe \cite{gangopadhyay2024uncle} proposes a benchmark and adapts existing methods \cite{kirkpatrick2017overcoming, li2017learning, rolnick2019experience} for unsupervised continual depth completion. To the best of our knowledge, we propose the first continual learning method for unsupervised depth completion and are the first to explore prototypes in this context.

\begin{figure*}[t]
  \centering
  \includegraphics[width=0.96\linewidth]{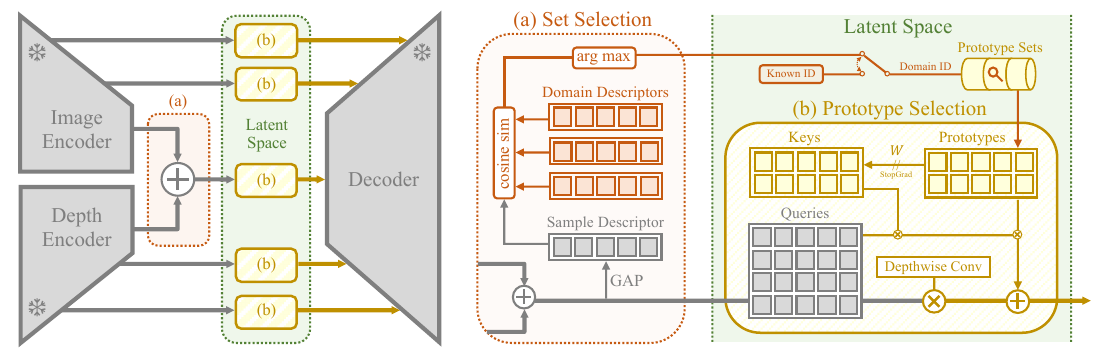}
  \vspace{-1mm}
  \caption{\textbf{Overview of ProtoDepth.} (a) 
  In the agnostic setting, a prototype set is selected by maximizing the cosine similarity between an input sample descriptor and the learned domain descriptors. In the incremental setting, the domain identity is known. (b)
  At inference, the similarity between the frozen queries and the keys of the selected prototype set determines how the learned prototypes contribute as local (additive) biases to the latent features. Additionally, a global (multiplicative) bias is applied using a $1\times1$ depthwise convolution.}
  \label{fig:method_diagram}
  \vspace{-3mm}
\end{figure*}

\section{Preliminaries}

\textbf{Unsupervised Depth Completion.} Assuming we are given an RGB image \( I : \Omega \subset \mathbb{R}^2 \rightarrow \mathbb{R}^3 \) and its associated sparse depth map \( z : \Omega \rightarrow \mathbb{R}_+ \) obtained by projecting the sparse point cloud onto the image plane, we wish to train a depth completion model \( f_\theta \) to predict the dense depth map \( \hat{d} \) in an unsupervised manner (i.e., without access to ground-truth depth). Unsupervised depth completion models~\cite{ma2019self, wong2020unsupervised, wong2021learning, wong2021unsupervised} typically minimize a loss function in the form of \cref{eqn:objective_function}, which comprises a linear combination of three terms:
\begin{equation}
    \label{eqn:objective_function}
    \mathcal{L} = w_{ph}\ell_{ph}+w_{sz}\ell_{sz}+w_{sm}\ell_{sm},
\end{equation}
where $\ell_{ph}$ denotes photometric consistency, $\ell_{sz}$ sparse depth consistency, and $\ell_{sm}$ a local smoothness regularizer.

\textit{Photometric Consistency} term leverages image reconstruction as the training signal. Specifically, given an image $I_t$ at time $t$, its reconstruction $\hat{I}_{t\tau}$ from a temporally adjacent image $I_{\tau}$ at time $\tau$ for $\tau \in \{t-1, t+1\}$ is given by 
\begin{equation}
\label{eqn:image_reconstruction}
    \hat{I}_{t\tau}(x, \hat{d}, g_{\tau t}) = I_{\tau} \big( \pi  g_{\tau t} K^{-1} \bar{x} \hat d(x) \big),
\end{equation}

where \( \bar{x} = [x^\top, 1]^\top \) is the homogeneous coordinates of \( x \in \Omega \), $K$ is the camera intrinsic calibration matrix, $g_{\tau t} \in SE(3)$ is the estimated relative camera pose matrix from time $t$ to $\tau$, and $\pi$ is the canonical perspective projection matrix. Given $I_t$ and its reconstruction $\hat{I}_{t\tau}$, the photometric consistency loss measures the $L1$ difference and structural similarity (\texttt{SSIM}~\cite{wang2004image}) between $I_t$ and $\hat{I}_{t\tau}$:
\begin{equation}
\begin{aligned}
  	\ell_{ph} = \frac{1}{|\Omega|} 
  	    \sum_{\tau\in T} \sum_{x \in \Omega}  
  	        &w_{co}| \hat{I}_{t\tau}(x)-I(x)| + \\ 
  	        &w_{st}\big(1 - \texttt{SSIM}(\hat{I}_{t\tau}(x), I(x))\big).
\label{eqn:photometric_consistency_loss}
\end{aligned}
\end{equation}

\textit{Sparse Depth Consistency.} However, photometric reconstruction recovers depth only up to an unknown scale. To ground predictions to a metric scale, we minimize an $L1$ loss between the predicted depth $\hat{d}$ and sparse depth $z$ for $x \in \Omega$ where points exist as denoted by $M : \Omega \mapsto \{0, 1\}$:
\begin{equation}
\label{eqn:sparse_depth_consistency_loss}
  	\ell_{sz} = \frac{1}{|\Omega|} 
  	    \sum_{x \in \Omega} 
  	        | M(x) \cdot (\hat{d}(x) - z(x))|. 
\end{equation}

\textit{Local Smoothness.} To address ambiguities in regions where the predicted depth is not constrained by photometric or sparse depth reconstruction terms, we rely on a regularizer that enforces local smoothness in predictions by applying an $L1$ penalty on the depth gradients in both the $x$-direction ($\partial_X$) and $y$-direction ($\partial_Y$). To allow for depth discontinuities along object boundaries, these penalties are weighted by their corresponding image gradients, $\lambda_{X} = e^{-|\partial_{X}I_{t}(x)|}$ and $\lambda_{Y} = e^{-|\partial_{Y}I_{t}(x)|}$. Larger image gradients result in smaller weights, allowing for sharp transitions in depth along edges:
\begin{equation}
\label{eqn:local_smoothness_loss}
  	\ell_{sm} = \frac{1}{|\Omega|}
  	    \sum_{x \in \Omega} 
      	    \lambda_{X}(x)|\partial_{X}\hat{d}(x)|+
      	    \lambda_{Y}(x)|\partial_{Y}\hat{d}(x)|.
\end{equation}

\textbf{Unsupervised Continual Depth Completion.} For continual learning, we consider a task sequence of domains $\mathcal{D}_1, \mathcal{D}_2, \cdots, \mathcal{D}_T$. Starting with a depth completion model $f_\theta$ pretrained on the initial dataset $\mathcal{D}_1$, we aim to incrementally adapt $f_\theta$ to each subsequent dataset $\mathcal{D}_2, \cdots, \mathcal{D}_k, \cdots, \mathcal{D}_T$. The key challenge is to learn the data distribution of each new dataset $\mathcal{D}_k$ without ``forgetting,'' as measured by performance degradation on previously learned datasets $\mathcal{D}_{j<k}$. We denote each dataset as $\mathcal{D}_k = \{ ( I_{k}^{(i)}, z_{k}^{(i)}, K_{k}^{(i)} ) \}_{i=1}^{n_k}$, which comprises $n_k$ training samples of image, sparse depth, and calibration, with no ground-truth depth. We assume that the relative camera pose $g_{t\tau}$ is given, or is estimated; if estimated by a pose network, we allow it to forget as we focus on continual learning for depth completion, and not pose estimation, which is outside the focus of this work.



\begin{table*}[t]
\scriptsize
\centering
\setlength\tabcolsep{2pt}
\resizebox{1.0\textwidth}{!}{%
\begin{tabular}{c l c c c c@{\hspace{7pt}} c c c c@{\hspace{7pt}} c c c c}
    \toprule
    & & \multicolumn{4}{c}{Average Forgetting (\%)} & \multicolumn{4}{c}{Average Performance (mm)\hspace*{6pt}} & \multicolumn{4}{c}{SPTO (mm)} \\
    \midrule
    Model & Method & MAE & RMSE & iMAE & iRMSE & MAE & RMSE & iMAE & iRMSE & MAE & RMSE & iMAE & iRMSE \\ 
    \midrule
    \multirow{6}*{VOICED}
    & Finetuned
    & 8.828 & 6.131 & 6.951 & 7.042 & 63.352 & 125.28 & 15.461 & 35.053& 52.453 & 108.434 & 15.360 & 35.357 
    \\
    & EWC~\cite{kirkpatrick2017overcoming}
    & 9.439 & 8.014 & 5.183 & 6.174 & 63.787 & 126.706 & 15.229 & 34.367 & 53.614 & 110.956 & 15.091 & 34.039
    \\
    & LwF~\cite{li2017learning}
    & 8.591 & 8.456 & 9.613 & 21.774 & 65.135 & 126.968 & 16.221 & 38.002 & 53.517 & 108.845 & 15.402 & 34.729
    \\
    & Replay~\cite{rolnick2019experience}
    & 6.154 & 4.688 & 9.471 & 11.713 & 64.305 & 126.714 & 16.373 & 36.729 & 54.326 & 112.218 & 16.640 & 37.671
    \\
    & \textit{\methodabbr-A}
    & \underline{2.439} & \underline{3.598} & \underline{4.630} & \underline{4.519}  & \underline{56.971} & \underline{118.132} & \textbf{13.554} & \underline{30.554} & \underline{47.367} & \underline{103.015} & \textbf{13.517} & \textbf{31.623}
    \\
    & \textit{\methodabbr}
    & \textbf{0.000} & \textbf{0.000} & \textbf{0.000} & \textbf{0.000} & \textbf{56.359} & \textbf{115.153} & \underline{13.589} & \textbf{30.332}& \textbf{46.934} & \textbf{101.326} & \underline{13.684} & \underline{31.925} 
    \\
    \midrule
    \multirow{6}*{FusionNet} 
    & Finetuned
    & 24.928 & 9.775 & 32.333 & 16.799  & 66.523 & 130.142 & 15.829 & 33.881& 54.252 & 110.666 & 15.317 & 33.726
    \\
    & EWC~\cite{kirkpatrick2017overcoming}
    & 11.256 & 8.782 & 17.944 & 17.847  & 64.487 & 130.890 & 15.264 & 34.203& 51.345 & 109.223 & 14.276 & 32.781
    \\
    & LwF~\cite{li2017learning}
    & 6.863 & 2.865 & 7.336 & 1.939& 61.204 & 123.573 & 14.075 & 30.879 & 50.159 & 106.386 & 13.879 & 31.608 
    \\
    & Replay~\cite{rolnick2019experience}
    & 5.702 & 2.862 & 12.196 & 11.186  & 61.467 & 125.587 & 14.750 & 33.279& 50.273 & 108.608 & 14.351 & 33.658
    \\
    & \textit{\methodabbr-A}
    & \underline{1.282} & \underline{0.686} & \underline{1.304} & \underline{0.446}  & \underline{57.742} & \underline{119.988} & \textbf{13.274} & \underline{30.139}& \underline{47.674} & \underline{104.349} & \underline{13.128} & \underline{31.058}
    \\
    & \textit{\methodabbr}
    & \textbf{0.000} & \textbf{0.000} & \textbf{0.000} & \textbf{0.000}  & \textbf{57.486} & \textbf{119.168} & \underline{13.323} & \textbf{29.936}& \textbf{47.335} & \textbf{102.845} & \textbf{13.091} & \textbf{30.474}
    \\
    \midrule
    \multirow{6}*{KBNet}
    & Finetuned
    & 16.080 & 15.463 & 8.188 & 9.170 
    
    & 58.577 & 124.606 & 13.474 & 31.409& 47.890 & 105.807 & \underline{13.266} & 31.742
    \\
    & EWC~\cite{kirkpatrick2017overcoming}
    & 14.915 & 11.878 & 10.398 & 5.640
    & 57.414 & 122.075 & 13.741 & 31.552& 48.031 & 106.661 & 14.129 & 33.096
    \\
    & LwF~\cite{li2017learning}
    & 9.717 & 6.324 & 6.168 & 5.254
    & 57.511 & 119.093 & 14.119 & 32.165& 47.154 & 103.164 & 14.304 & 33.838
    \\ 
    & Replay~\cite{rolnick2019experience}
    & 7.200 & 4.819 & 9.202 & 9.539
    & 56.208 & 117.848 & 13.983 & 32.341& 46.700 & 103.631 & 13.844 & 33.326
    \\
    & \textit{\methodabbr-A}
    & \underline{3.204} & \underline{1.304} & \underline{4.911} & \underline{2.943}
    & \underline{54.254} & \underline{115.548} & \underline{13.201} & \underline{30.499}& \underline{45.264} & \underline{101.097} & 13.281 & \underline{31.718}
    \\
    & \textit{\methodabbr}
    & \textbf{0.000} & \textbf{0.000} & \textbf{0.000} & \textbf{0.000} 
    & \textbf{52.497} & \textbf{113.548} & \textbf{12.845} & \textbf{29.990}& \textbf{44.092} & \textbf{99.788} & \textbf{13.081} & \textbf{31.503}
    \\
    \midrule
\end{tabular}
}
\vspace{-3mm}
\caption{\textbf{Quantitative results} on \textbf{indoor} datasets. Models are initially trained on NYUv2 and continually trained on ScanNet, then VOID. \textbf{Bold} indicates the best performance, while \underline{underline} indicates the second-best performance. Baseline results are obtained from UnCLe~\cite{gangopadhyay2024uncle}.
}
\vspace{-3mm}
\label{tab:indoor}
\end{table*}

\section{Method}

We present \textbf{\methodabbr}, a novel approach for unsupervised continual depth completion that mitigates catastrophic forgetting by leveraging prototype sets as selective biases. Given a pretrained depth completion model $f_\theta$, which we freeze to prevent any forgetting, we adapt it to new datasets by deploying lightweight, domain-specific prototype sets that learn to selectively bias the latent features; note that this only adds minimal additional parameters per dataset or domain. 
Our method is applicable to the \emph{domain-incremental} (``incremental'') setting, where dataset identity is known at test-time, and the more challenging \emph{domain-agnostic} (``agnostic'') setting, where the test-time domain identity is unknown, through a proposed prototype set selection mechanism (see \cref{sec:prototype_set_selection}).

\subsection{Prototype Learning} \label{sec:prototypes}


To enable the model to adapt to new datasets without forgetting, we learn layer-specific \emph{prototype sets} for each dataset that serve as multiplicative (global) and additive (local) biases in the latent feature space. For simplicity, we consider an input sample from a single dataset $\mathcal{D}_k$ at a single layer $l$, which is encoded into the latent features $X \in \mathbb{R}^{h\times w \times c}$. We assume a linear transformation from the learned latent space of $\mathcal{D}_1$ to that of $\mathcal{D}_k$; hence, we formulate the adaptation as 
\begin{equation}
\label{eqn:enhanced_features}
    \hat{X} = A \odot X + B,
\end{equation}
where $\odot$ denotes a (broadcasted) Hadamard product between the global prototype $A \in \mathbb{R}^c$ and the features $X$; $B$ is an additive bias constructed from a set of local prototypes $P$.

To this end, we flatten the latent features $X$ to get $Q \in \mathbb{R}^{(h\times w) \times c}$. Since the model $f_\theta$ is frozen, $Q$ serves as a set of $h\times w$ deterministic \emph{queries}, where each query is a $c$-dimensional vector. We introduce $N$ learnable additive prototypes $P = [p_{1}, p_{2}, \cdots, p_{N}]^{\top} \in \mathbb{R}^{N \times c}$, where each $p_{i}$ is a $c$-dimensional vector representing a ``prototypical'' local feature of the dataset. To learn the \emph{keys} associated with each prototype, we define a projection matrix $W$ that learns to map the prototypes $P$ back into the query space, i.e., the latent feature space. This yields $K \in \mathbb{R}^{N \times c}$, where
\begin{equation}
\label{eqn:key_projection}
    K = \text{StopGrad}(P)\times W.
\end{equation}
StopGrad (stop gradient) facilitates decoupled optimization, enabling prototypes to learn appropriate additive biases while keys learn to assign relevant prototypes to queries.
We compute the similarity scores between the queries $Q$ and the keys $K$ using scaled dot-product attention~\cite{vaswani2017attention}. To obtain the additive bias $b \in \mathbb{R}^{(h\times w) \times c}$, the scores are used to compute a convex combination of prototypes $P$:
\begin{equation}
\label{eqn:attention}
    b = \text{softmax}\left(Q \times K^{\top} \middle/ \sqrt{c}\,\right) \times P.
\end{equation}




We reshape $b$ back to the spatial dimensions of $X$ to obtain the local additive bias $B \in \mathbb{R}^{h\times w\times c}$. To model the global transformation, we learn a $c$-dimensional multiplicative prototype $A$, applied element-wise as $A \odot X$, which can be efficiently implemented as a $1 \times 1$ depthwise convolution. The result is further adapted to the new dataset distribution, still without altering the original model parameters, by incorporating the local domain-specific transformation $B$ as an additive bias, 
i.e., $\hat{X} = A \odot X + B$. As $f_\theta$ is frozen and a new prototype set (local and global prototypes $P_k$ and $A_k$, and projection matrix $W_k$) is learned for each dataset $\mathcal{D}_k$, this naturally facilitates continual learning and ensures no forgetting in the incremental setting, where the prototype set corresponding to the domain identity is selected. We further extend this to the agnostic setting in \cref{sec:prototype_set_selection}.



\subsection{\methodabbr Architecture}

Current unsupervised depth completion models~\cite{wong2021unsupervised, wong2021learning,wong2020unsupervised} adopt an encoder-decoder CNN architecture, which consists of separate image and sparse depth encoders with skip connections to the decoder. We refer to the bottleneck and the skip connections as the latent space layers (see \cref{fig:method_diagram}).

To extend the prototype mechanism (\cref{sec:prototypes}) across multiple layers, for each new dataset $D_k$, we introduce a prototype \textit{set} of local and global prototypes $P^{(l)}$ and $A^{(l)}$, and projection matrix $W^{(l)}$ for each layer $l$ in the latent space. For each new dataset, the latent feature adaptation (\cref{eqn:attention,eqn:enhanced_features}) is applied independently to each layer $l$.

As different modalities in multimodal tasks (e.g., RGB image and sparse depth map in depth completion) may experience varying degrees of covariate shift across domains, we propose to deploy a different number of prototypes $N^{(I)}$ and $N^{(z)}$ for the RGB image and sparse depth modalities, respectively. Based on the observation that RGB images undergo a larger covariate shift than sparse depth~\cite{park2024testtime}, we choose $N^{(I)} > N^{(z)}$ to capture their prototypical features; this choice reduces the parameter overhead.

The proposed prototype-based continual learning mechanism operates on the latent feature space and does not depend on the specific architecture of the model. This architecture-agnostic flexibility stems from the fact that our queries $Q\in\mathbb{R}^{(h\times w)\times c}$ mirror the general structure of latent features across commonly used model architectures, where $h\times w$ can be replaced by the number of tokens $n$ in the case of transformers \cite{vaswani2017attention}. Thus, it can be applied generically to models with latent feature representations~\cite{lao2024sub, xia2023quadric}, providing a general framework for mitigating catastrophic forgetting across various tasks and modalities.

\begin{table*}[t]
\scriptsize
\centering
\setlength\tabcolsep{2pt}
\resizebox{1.0\textwidth}{!}{%
\begin{tabular}{c l c c c c@{\hspace{7pt}} c c c c@{\hspace{7pt}} c c c c}
    \toprule
    & & \multicolumn{4}{c}{Average Forgetting (\%)} & \multicolumn{4}{c}{Average Performance (mm)\hspace*{6pt}} & \multicolumn{4}{c}{SPTO (mm)} \\
    \midrule
    Model & Method & MAE & RMSE & iMAE & iRMSE & MAE & RMSE & iMAE & iRMSE & MAE & RMSE & iMAE & iRMSE \\ 
    \midrule
    \multirow{6}*{VOICED}
    & Finetuned
    & 499.598 & 162.188 & 467.472 & 208.693  & 1620.429 & 3072.129 & 4.040 & 6.144& 914.223 & 2993.228 & 1.955 & 4.503
    \\
    & EWC~\cite{kirkpatrick2017overcoming}
    & 555.925 & 190.152 & 540.109 & 247.943  & 1796.300 & 3346.057 & 4.490 & 6.685& 962.937 & 3209.759 & 1.962 & 4.739
    \\
    & LwF~\cite{li2017learning}
    & 631.119 & 221.535 & 524.976 & 233.758  & 1973.972 & 3612.700 & 4.533 & 6.648& 985.995 & 3236.244 & 2.062 & 4.722
    \\
    & Replay~\cite{rolnick2019experience}
    & 17.241 & 4.050 & 16.662 & 5.478 & 524.114 & 1875.897 & 1.333 & 3.359& 618.668 & 2366.577 & 1.292 & 3.348 
    \\
    & \textit{\methodabbr-A}
    & \underline{2.427} & \underline{2.863} & \underline{2.079} & \underline{2.153} & \underline{458.520} & \underline{1832.690} & \underline{1.133} & \underline{3.213}& \underline{548.240} & \underline{2294.399} & \underline{1.080} & \underline{3.159} 
    \\
    & \textit{\methodabbr}
    & \textbf{0.000} & \textbf{0.000} & \textbf{0.000} & \textbf{0.000}  & \textbf{445.419} & \textbf{1804.158} & \textbf{1.106} & \textbf{3.169}& \textbf{531.689} & \textbf{2262.943} & \textbf{1.043} & \textbf{3.110}
    \\
    \midrule
    \multirow{6}*{FusionNet} 
    & Finetuned
    & 11.336 & 8.435 & 17.447 & 17.991  & 437.730 & 1785.212 & 1.193 & 3.724& 501.362 & 2138.422 & 1.111 & 3.978
    \\
    & EWC~\cite{kirkpatrick2017overcoming}
    & 21.006 & 10.494 & 20.431 & 16.535 & 431.440 & 1760.460 & 1.144 & 3.181 & 486.170 & 2117.030 & 1.029 & 2.986
    \\
    & LwF~\cite{li2017learning}
    & 12.368 & 5.202 & 13.593 & 13.117 & 442.878 & 1759.202 & 1.178 & 3.352 & 526.528 & 2168.961 & 1.156 & 3.451
    \\
    & Replay~\cite{rolnick2019experience}
    & 8.290 & 11.134 & 2.769 & 7.975  & 419.044 & 1774.361 & 1.044 & 3.032& 479.168 & 2122.997 & 0.966 & 2.906
    \\
    & \textit{\methodabbr-A}
    & \underline{2.200} & \underline{2.282} & \underline{2.602} & \underline{7.203}  & \underline{404.956} & \underline{1702.945} & \underline{1.041} & \underline{3.028}& \underline{464.976} & \underline{2052.413} & \underline{0.952} & \underline{2.864}
    \\
    & \textit{\methodabbr}
    & \textbf{0.000} & \textbf{0.000} & \textbf{0.000} & \textbf{0.000}  & \textbf{400.888} & \textbf{1683.202} & \textbf{1.022} & \textbf{2.899}& \textbf{461.043} & \textbf{2048.942} & \textbf{0.932} & \textbf{2.792}
    \\
    \midrule
    \multirow{6}*{KBNet}
    & Finetuned
    & 27.153 & 18.208 & 52.969 & 33.370 & 469.658 & 1943.259 & 1.338 & 3.683& 541.383 & 2411.169 & 1.144 & 3.505 
    \\
    & EWC~\cite{kirkpatrick2017overcoming}
    & 23.517 & 8.583 & 30.077 & 18.991  & 456.828 & 1806.761 & 1.221 & 3.321& 526.366 & 2210.424 & 1.133 & 3.158
    \\
    & LwF~\cite{li2017learning}
    & 21.184 & 4.049 & 43.500 & 19.951& 460.097 & 1749.734 & 1.362 & 3.555 & 541.932 & 2142.999 & 1.359 & 3.731 
    \\ 
    & Replay~\cite{rolnick2019experience}
    & 25.423 & 29.303 & 6.362 & 7.274 & 454.896 & 1935.667 & 1.102 & 3.203& 525.696 & 2318.363 & 1.094 & 3.246 
    \\
    & \textit{\methodabbr-A}
    & \underline{4.513} & \underline{3.100} & \underline{2.960} & \underline{1.878}  & \underline{409.903} & \underline{1730.720} & \underline{1.045} & \underline{3.044}& \underline{478.790} & \underline{2138.347} & \underline{1.008} & \underline{3.066}
    \\
    & \textit{\methodabbr}
    & \textbf{0.000} & \textbf{0.000} & \textbf{0.000} & \textbf{0.000}  & \textbf{401.075} & \textbf{1710.074} & \textbf{1.029} & \textbf{2.993}& \textbf{471.437} & \textbf{2125.957} & \textbf{0.996} & \textbf{3.015}
    \\
    \midrule
\end{tabular}
}
\vspace{-3mm}
\caption{\textbf{Quantitative results} on \textbf{outdoor} datasets. Models are initially trained on KITTI and continually trained on Waymo, then VKITTI. \textbf{Bold} indicates the best performance, while \underline{underline} indicates the second-best performance. Baseline results are obtained from UnCLe~\cite{gangopadhyay2024uncle}.
}
\vspace{-3mm}
\label{tab:outdoor}
\end{table*}

\subsection{Prototype Set Selection}
\label{sec:prototype_set_selection}

As the prototypes are learned for a specific domain, we cannot easily select the appropriate prototype set for inference if the test-time domain identity is withheld, i.e., in the domain-agnostic setting. To address this challenging scenario, we introduce a prototype set selection mechanism that chooses the most relevant prototype set for a given input.

During training, we introduce a \textit{domain descriptor} $r_k \in \mathbb{R}^c$ for each dataset $\mathcal{D}_k$, which adds negligible overhead in terms of number of parameters. For an input from $\mathcal{D}_k$, we obtain a \emph{sample descriptor} $s_k \in \mathbb{R}^c$ by applying global average pooling (GAP) to the bottleneck latent features (with channel dimension $c$) before applying the prototype set. Importantly, since both encoders are always frozen during continual training, $s_k$ is a deterministic mapping of the input.

For each new dataset $\mathcal{D}_k$, we deploy a new domain descriptor $r_k$ and freeze all existing learned domain descriptors. The deployed domain descriptor $r_k$ is trained by minimizing cosine distance between itself and sample descriptors $s_k$ for $\mathcal{D}_k$, while maximizing the cosine distance to all other learned domain descriptors $\left\{r_{j\neq k}\right\}$. 
This naturally yields domain descriptors that are discriminative across datasets, allowing us to use the projection of sample descriptors onto domain descriptors as a prototype set selection mechanism. To this end, we propose to minimize an additional objective:
\begin{equation}
\label{eqn:dataset_representation_loss}
  	\ell_{dr} =
            1 - (\frac{s_k}{||s_k||}\cdot\frac{r_k}{||r_k||})
            + \frac{1}{w_{jk}} \sum_{j \neq k} (\frac{r_j}{||r_j||}\cdot\frac{r_k}{||r_k||}),
\end{equation}
where $||\cdot||$ denotes the $L2$-norm and $w_{jk} \propto |j \neq k|$ is a tunable normalization constant. As the previously learned domain descriptors are frozen, their alignment to their respective datasets or domains is preserved, allowing us to continually learn new domain descriptors that can distinguish new datasets. \cref{eqn:dataset_representation_loss} is incorporated into \cref{eqn:objective_function} as an additional term weighted by $w_{dr}$ to yield a novel loss function for training of unsupervised continual depth completion models in the agnostic setting:
\begin{equation}
    \label{eqn:overall_objective_function}
    \mathcal{L} = w_{ph}\ell_{ph}+w_{sz}\ell_{sz}+w_{sm}\ell_{sm}+w_{dr}\ell_{dr}.
\end{equation}

At test-time, we compute the sample descriptor $s$ for an input without dataset identity and select the domain descriptor $r_{k^*}$ that maximizes cosine similarity with $s$:
\begin{equation}
    \label{eqn:representation_selection}
    k^* = \arg\max_{k}\, (\frac{s}{||s||}\cdot\frac{r_k}{||r_k||}).
\end{equation}

For each latent space layer, we use the prototype set corresponding to the selected domain descriptor. While this does not eliminate forgetting due to the evolving set of domain descriptors and possible overlap between domains, it does minimize forgetting as each prototype set is learned independently for each dataset, but can still be selectively used for inference without knowing the test-time dataset identity. 
The trade-off is shown in \cref{tab:indoor,tab:outdoor} (\methodabbr-A) where we incur forgetting in exchange for the flexibility to support both the incremental and agnostic settings.

\begin{figure*}[t]
  \centering
  \includegraphics[width=1.0\linewidth]{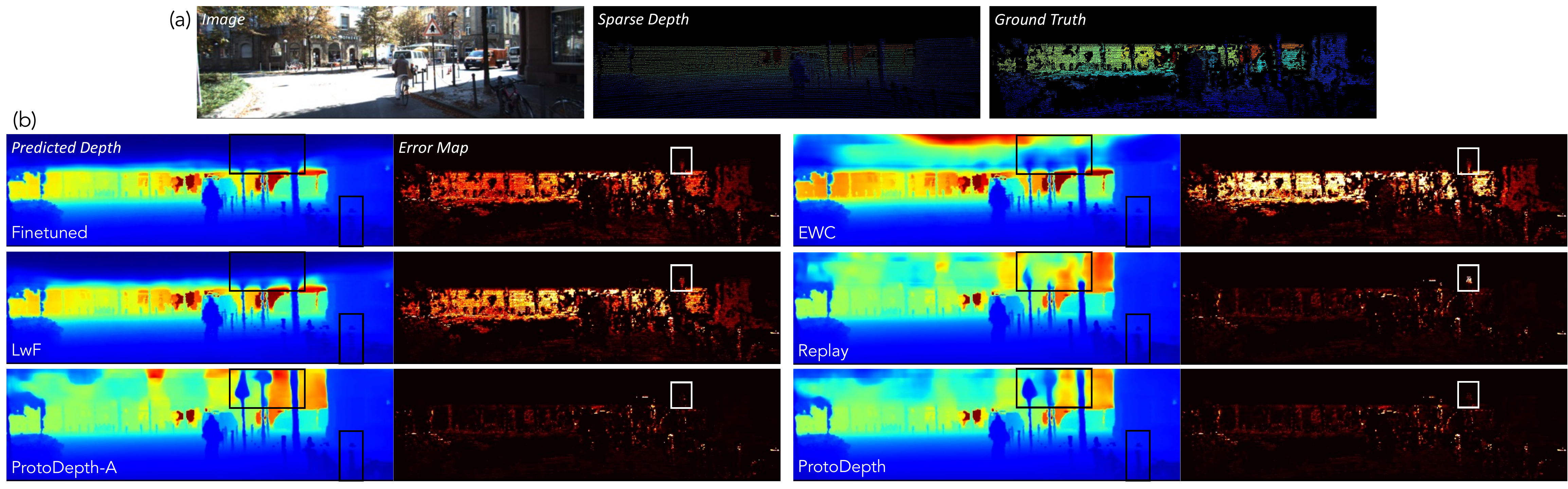}
  \vspace{-6mm}
  \caption{\textbf{Qualitative comparison} of \methodabbr and baseline methods using VOICED on \textbf{KITTI} after continual training on \textbf{Waymo}. (a) Input sample from KITTI, (b) Baseline methods exhibit significant forgetting, particularly for small-surface-area objects (e.g., street signs and lamp posts) where sparse depth is limited, and photometric priors from KITTI are critical. In contrast, \methodabbr produces high-fidelity depth predictions, effectively mitigating forgetting despite the large domain gap between KITTI and Waymo.}
  \label{fig:qualitative}
  \vspace{-3mm}
\end{figure*}

\section{Experiments}



\hspace*{1pc}\textbf{Datasets.} \underline{Indoor} dataset sequence: \emph{NYUv2}~\cite{silberman2012indoor} contains household, office, and commercial scenes captured with a Microsoft Kinect; \emph{ScanNet}~\cite{dai2017scannet} is a diverse, large-scale dataset captured using a Structure Sensor; \emph{VOID}~\cite{wong2020unsupervised} contains laboratory, classroom, and garden scenes captured using XIVO.
\underline{Outdoor} dataset sequence: \emph{KITTI}~\cite{uhrig2017sparsity} is a daytime autonomous driving benchmark captured using a Velodyne LiDAR sensor; \emph{Waymo}~\cite{sun2020scalability} contains road scenes with a wide variety of driving conditions;
\emph{VKITTI}~\cite{gaidon2016virtual} is a synthetic dataset that replicates and augments KITTI scenes.


\textbf{Models.} We evaluate using three recent unsupervised depth completion models in the continual learning setting: VOICED~\cite{wong2020unsupervised}, FusionNet~\cite{wong2021learning}, and KBNet~\cite{wong2021unsupervised}. 

\textbf{Baseline Methods.} We compare \methodabbr against EWC~\cite{kirkpatrick2017overcoming}, LwF~\cite{li2017learning}, and Experience Replay (``Replay'')~\cite{rolnick2019experience} as milestone works of their respective class of continual learning approaches. 
We include full finetuning (``Finetuned'') as a baseline of performance with no continual learning strategy. All baseline methods achieve identical performance in the incremental and agnostic settings.


\textbf{Evaluation Metrics} are computed across four standard depth completion metrics (MAE, RMSE, iMAE, iRMSE). We define the following evaluation metrics in terms of $a_j^k$, denoting any one of the four depth completion metrics on dataset $\mathcal{D}_{j}$ after training on $\mathcal{D}_{k}$. Given $T$ total datasets:

\emph{Average Forgetting} ($\bar{F}$) is the scale-invariant mean of how much performance on previous datasets $\mathcal{D}_{j<k}$ deteriorates (i.e., increases in \%) after training on each new $\mathcal{D}_{k}$:

\begin{equation}
\label{eqn:average_forgetting}
    \bar{F} = \frac{2}{T(T-1)}\, \sum_{k=1}^T\, \sum_{j<k}\, \frac{a_j^k - a_j^j}{a_j^j}.
\end{equation}

\emph{Average Performance} ($\bar{\mu}$) is the mean of performance on all seen datasets $\mathcal{D}_{j\leq k}$ after training on each new $\mathcal{D}_{k}$:

\begin{equation}
\label{eqn:average_performance}
    \bar{\mu} = \frac{2}{T(T+1)}\, \sum_{k=1}^T\, \sum_{j\leq k}\, a_j^k.
\end{equation}

\emph{Stability-Plasticity Trade-off} (SPTO) captures the balance between retaining learned knowledge (stability) and adapting to new domains (plasticity) as a harmonic mean:

\begin{equation}
\label{eqn:spto}
    \text{SPTO} = \frac{2\times S\times P}{S+P},\,
\begin{cases}
S = \sum_{k=1}^T a_k^T \\[0.25em]
P = \sum_{k=1}^T a_k^k\,,
\end{cases}
\end{equation}

where $S$ is performance across all datasets after completing training on the dataset sequence, and $P$ is performance on each new dataset after training on it for the first time.

\begin{table*}[t]
\scriptsize
\centering
\setlength\tabcolsep{2pt}
\resizebox{1.0\textwidth}{!}{%
\begin{tabular}{c c c c@{\hspace{8pt}} c c c c@{\hspace{8pt}} c c c c}
    \toprule
    & & & & \multicolumn{4}{c}{ScanNet} & \multicolumn{4}{c}{VOID} \\
    \midrule
    Method & $N^{(I)}$ & $N^{(z)}$ & \# Params & MAE & RMSE & iMAE & iRMSE  & MAE & RMSE & iMAE & iRMSE \\ 
    \midrule
    Pretrained & - & - & 0M (0\%)
    & 4114.04 & 4626.00 & 390.78 & 447.31
    & 42.94 & 106.39 & 29.26 & 64.04
    \\
    \midrule
    \multirow{5.5}*{\methodabbr} & 
    1 & 1 & 0.24M (3.5\%)
    & 19.68\tiny{$\pm$0.68} & 60.10\tiny{$\pm$0.71} & 10.44\tiny{$\pm$0.48} & 27.86\tiny{$\pm$0.74}
    & 37.81\tiny{$\pm$0.52} & 93.72\tiny{$\pm$0.90} & 22.82\tiny{$\pm$0.27} & 52.58\tiny{$\pm$0.32}
    \vspace{0.5ex}
    \\
    & 5 & 5 & 0.25M (3.6\%)
    & 16.49\tiny{$\pm$0.15} & 57.51\tiny{$\pm$0.19} & 6.84\tiny{$\pm$0.03} & 22.07\tiny{$\pm$0.03}
    & 34.02\tiny{$\pm$0.25} & 87.72\tiny{$\pm$0.43} & 17.92\tiny{$\pm$0.14} & 43.95\tiny{$\pm$0.28}
    \vspace{0.5ex}
    \\
    & 10 & 5 & 0.25M (3.6\%)
    & \textbf{14.59}\tiny{$\pm$0.17} & \textbf{42.20}\tiny{$\pm$0.09} & \textbf{5.57}\tiny{$\pm$0.14} & \textbf{17.10}\tiny{$\pm$0.15}
    & \textbf{33.63}\tiny{$\pm$0.23} & \textbf{87.30}\tiny{$\pm$0.57} & \textbf{17.55}\tiny{$\pm$0.32} & \textbf{43.24}\tiny{$\pm$0.63}
    \vspace{0.5ex}
    \\
    & 10 & 10 & 0.25M (3.7\%)
    & 15.25\tiny{$\pm$0.39} & 43.31\tiny{$\pm$0.72} & 5.85\tiny{$\pm$0.22} & 17.57\tiny{$\pm$0.38}
    & 34.39\tiny{$\pm$0.82} & 88.73\tiny{$\pm$1.83} & 18.49\tiny{$\pm$0.85} & 45.23\tiny{$\pm$1.60}
    \vspace{0.5ex}
    \\
    & 100 & 100 & 0.38M (5.5\%)
    & 16.35\tiny{$\pm$0.37} & 47.61\tiny{$\pm$0.16} & 5.90\tiny{$\pm$0.17} & 20.11\tiny{$\pm$0.19}
    & 34.29\tiny{$\pm$0.58} & 88.22\tiny{$\pm$0.86} & 18.16\tiny{$\pm$0.62} & 44.51\tiny{$\pm$1.07}
    \\
    \midrule
\end{tabular}
}
\vspace{-3mm}
\caption{\textbf{Sensitivity study} of prototype set sizes ($N^{(I)}$ and $N^{(z)}$) on \methodabbr using KBNet for indoor datasets (ScanNet and VOID). KBNet is pretrained on the initial dataset (NYUv2). Parameter overhead is reported as a percentage of the full KBNet model's parameters. 
}
\vspace{-3mm}
\label{tab:indoor_set_sizes}
\end{table*}

\begin{table}[t]
\scriptsize
\centering
\setlength\tabcolsep{2pt}
\resizebox{\columnwidth}{!}
{%
\begin{tabular}{l@{\hspace{8pt}} c c@{\hspace{8pt}} c c}
    \toprule
    & \multicolumn{2}{c}{ScanNet\hspace*{7pt}} & \multicolumn{2}{c}{Waymo} \\
    \midrule
    Ablated Component & MAE & RMSE & MAE & RMSE\\
    \midrule 
    global prototypes & 18.12 & 58.91 & 505.01 & 1715.21 \\
    projection matrix $W$ & 17.59 & 57.61 & 495.05 & 1690.51 \\
    decoupled $K$ and $P$  & 16.36 & 45.10 & 491.59 & 1675.57 \\
    \midrule
    no ablations & 14.59 & 42.20 & 486.95 & 1664.18 \\
    \midrule
\end{tabular}
}
\vspace{-3mm}
\caption{\textbf{Ablation studies} using KBNet for indoor and outdoor.
}
\vspace{-3mm}
\label{tab:ablations}
\end{table}

\subsection{Main Results}

We compare our method, evaluated in both the incremental (\emph{\methodabbr}) and agnostic (\emph{\methodabbr-A}) settings, against baseline methods for the indoor dataset sequence in \cref{tab:indoor} and for the outdoor dataset sequence in \cref{tab:outdoor}. 

\textbf{Results in Incremental Setting.} In both indoor and outdoor settings, \methodabbr achieves a 100\% improvement in Average Forgetting compared to all baseline methods across all models and metrics. This is, of course, because \methodabbr exhibits zero forgetting as it freezes all model parameters and learns dataset-specific prototypes. For the indoor sequence, compared to the best baseline method, \methodabbr improves Average Performance by 5.15\% and SPTO by 6.59\%, averaged across all models and metrics. Similarly, for the outdoor sequence, we improve Average Performance by 6.88\% and SPTO by 6.94\%.

To demonstrate the reduced forgetting achieved by \methodabbr, we qualitatively compare against all baseline methods using VOICED on KITTI after continual training on Waymo (see \cref{fig:qualitative}). 
\methodabbr yields better depth predictions for small-surface-area objects with limited sparse depth measurements for which the model must rely on photometric priors learned from images. Unlike KITTI, which consists exclusively of daytime scenes, Waymo includes many evening and overcast scenes, introducing variations in lighting and pixel intensities. Additionally, Waymo was captured using a higher-resolution camera which causes objects to appear bigger in terms of number of pixels occupied. Due to this large distributional shift, the model forgets the \emph{projected} shapes of objects in KITTI after training on Waymo, even if the objects exist in both datasets. This forgetting is apparent in the highlighted street sign and lamp posts, where baseline methods struggle to accurately predict depth.


\textbf{Results in Agnostic Setting.} For the indoor sequence, compared to the best baseline method, \methodabbr-A improves Average Forgetting by 52.22\%, Average Performance by 4.26\%, and SPTO by 5.40\%, averaged across all models and metrics. Notably, \methodabbr-A outperforms \methodabbr in some metrics, meaning the model appropriately selects prototypes of different domains when there is domain overlap, thereby enhancing its generalization capabilities.

For the outdoor sequence, \methodabbr-A shows an average improvement of 53.21\% in Average Forgetting across all models and metrics. In contrast to the indoor sequence, \methodabbr-A does not outperform \methodabbr in any metric, likely due to the larger domain gaps between the outdoor datasets. Selecting prototypes from a different outdoor dataset is more likely to be erroneous, leading to performance degradation rather than generalization.

Furthermore, we refer back to \cref{fig:qualitative} (\methodabbr-A) for head-to-head comparison of our method against other baselines in the agnostic setting. The error maps for Finetuned, EWC, and LwF display significant errors, indicating substantial forgetting of previously learned information. While Replay yields an improved error map, it still experiences forgetting in small-surface-area objects. For example, Replay fails to reconstruct the upper portions of the highlighted street sign and lamp posts due to forgetting of learned photometric priors from KITTI, whereas \methodabbr-A recalls them from KITTI prototypes. Additionally, \methodabbr-A predicts the depth of the highlighted small fence poles with higher fidelity than the incoherent prediction of Replay.

\subsection{Design Choice Studies} \label{sec:ablations}

\hspace*{1pc}\textbf{Prototype Set Sizes.} We investigate the impact of varying the prototype set sizes (i.e., number of prototypes) for the image and sparse depth layers (denoted as $N^{(I)}$ and $N^{(z)}$, respectively) on the performance of our method. The set size experiments for the indoor sequence are shown in \cref{tab:indoor_set_sizes}, based on which we selected $N^{(I)} = 10, N^{(z)} = 5$ for the main experiments. 
Smaller set sizes 
perform worse as there is insufficient capacity to capture the diversity of features in each dataset. There is also performance degradation with larger set sizes; intuitively, unnecessary additional parameters may learn noise and cause overfitting. Notably, best performance is achieved when $N^{(I)} > N^{(z)}$, which can be attributed to the larger distributional shift between scenes in the image modality compared to the sparse depth modality~\cite{park2024testtime}. Since the bottleneck layer fuses both modalities, we use $N^{(I)}$ for the bottleneck layer prototypes. As a lower bound, we show that the frozen base model pretrained on NYUv2 (``Pretrained'') 
performs poorly, motivating the need for continual learning. 
We perform similar set size experiments for the outdoor dataset sequence (see Supp. Mat.), based on which we choose $N^{(I)} = 25, N^{(z)} = 10$.

\textbf{Ablations.} We assess the impact of the components of \methodabbr on both indoor (ScanNet) and outdoor (Waymo) in \cref{tab:ablations}. Removing the $1\times1$ depthwise convolutions results in performance degradation, demonstrating their effectiveness as lightweight global prototypes. Learning the keys $K$ independently from the prototypes $P$ without the projection matrix $W$ hurts performance, suggesting that the projection matrix effectively learns to map the prototypes into latent feature space, fulfilling the intended role of keys. Furthermore, performance decreases without the stop gradient operation on $P$ when computing $K$, indicating the importance of decoupled optimization of keys and prototypes.

\section{Discussion}

\methodabbr leverages prototypes as a mechanism for mitigating catastrophic forgetting.
While we demonstrate it on unsupervised depth completion, \methodabbr does not assume specific modalities and thus can be relevant to other multimodal problems \cite{xie2023sparsefusion, yang2024binding, zeng2024wordepth}.
Our promising results on both indoor and outdoor domains illustrate the potential for \methodabbr to enable unsupervised continual learning for multimodal 3D reconstruction. 
Our architecture-agnostic approach can also be extended to other tasks involving models that produce latent feature representations \cite{lao2024sub, zeng2024rsa}, offering a general framework for continual learning.

\textbf{Limitations.} \methodabbr relies on knowledge of dataset boundaries to instantiate new prototype sets, which may not be feasible in online training settings where there are no defined boundaries between domains. In the same vein, we do not consider scenarios where domain gaps between datasets are small or where there are significant distributional shifts within a dataset. Addressing these limitations would require mechanisms to dynamically detect domain shifts and instantiate new prototypes when appropriate.

\section*{Acknowledgments}

This work is supported by NSF-2112562 Athena AI Institute.

{
    \small
    \bibliographystyle{ieeenat_fullname}
    \bibliography{bibtex/depth_completion, bibtex/continual_learning, bibtex/method}
}


\clearpage   
\appendix

\maketitlesupplementary

\begin{figure}[b]
    \centering
    \includegraphics[width=0.9\linewidth, trim=50 20 50 30]{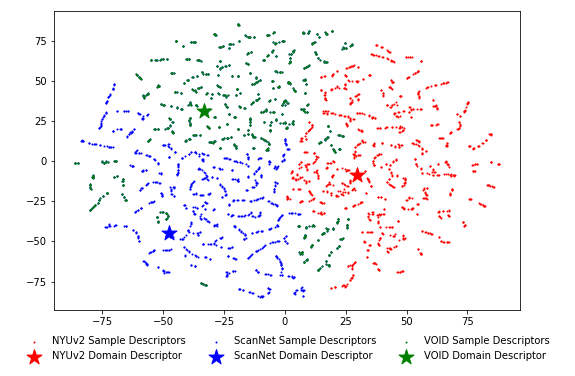}
    \vspace{2mm}
    \caption{\textbf{t-SNE plot} of KBNet sample descriptors for indoor validation datasets (NYUv2, ScanNet, VOID) and their respective domain descriptors learned during training in the agnostic setting. While most sample descriptors align closely with their respective domain descriptors, some overlap enables cross-domain generalization, improving performance in challenging scenarios.}
    
    \label{fig:t-sne}
\end{figure}

\section{Domain Descriptor Analysis}

To better understand the performance of ProtoDepth in the agnostic setting, we analyze the relationship between sample descriptors and learned domain descriptors using the t-SNE visualization shown in \cref{fig:t-sne}. This analysis is based on the KBNet model trained on the indoor dataset sequence, and it reveals insights into how ProtoDepth selects prototype sets during inference.

Each sample descriptor is computed deterministically using global average pooling (GAP) over the bottleneck features of the frozen model. Since the encoder layers are always frozen during training, the sample descriptors of a certain dataset are a lifelong deterministic function of the features present in that dataset. The domain descriptors, on the other hand, are learned during training to align with the sample descriptors of their respective datasets, enabling effective prototype set selection.

The visualization demonstrates that the majority of sample descriptors for each dataset cluster closely around their respective domain descriptors. This alignment confirms that the training process successfully associates each dataset with its corresponding descriptor at test-time, ensuring accurate prototype selection in the agnostic setting. However, it is noteworthy that some sample descriptors are closer to domain descriptors of other datasets. For example, non-negligible subsets of VOID sample descriptors appear to have higher affinity with the NYUv2 and ScanNet domain descriptors. This overlap introduces a degree of generalization, allowing the model to select prototypes from a different domain if they better align with the input sample's features.

This ability to adaptively select domain descriptors explains why ProtoDepth achieves superior performance in the agnostic setting than in the incremental setting for certain metrics. By relaxing the constraint of fixed domain identity during inference, the agnostic setting enables the model to exploit cross-domain generalization in cases where overlapping features exist between datasets. While this occurs in only a minority of scenarios, it underscores the utility of allowing the model to flexibly choose prototypes, particularly in instances where the distributional characteristics of one domain may overlap with those of another.

Most importantly, the t-SNE plot clearly illustrates that, despite the presence of some overlap, the domain descriptors remain sufficiently distinct to avoid significant performance degradation due to incorrect prototype selection. Instead, this overlap even facilitates generalization (see \cref{tab:generalization}), enabling the model to leverage features from neighboring domains to improve depth completion on difficult samples. This balance between dataset alignment and cross-domain generalization is central to ProtoDepth's ability to adapt to the challenging domain-agnostic setting.

\begin{table*}[t]
\scriptsize
\centering
\setlength\tabcolsep{2pt}
\resizebox{1.0\textwidth}{!}{%
\begin{tabular}{c l c c c c@{\hspace{7pt}} c c c c@{\hspace{7pt}} c c c c}
    \toprule
    \multicolumn{2}{l}{} & \multicolumn{4}{c}{Average Forgetting (\%)} & \multicolumn{4}{c}{Average Performance\hspace*{6pt}} & \multicolumn{4}{c}{SPTO} \\
    Setting & Method & MAE & RMSE & iMAE & iRMSE & MAE & RMSE & iMAE & iRMSE & MAE & RMSE & iMAE & iRMSE \\ [-0.5mm]
    \midrule
    \multirow{3}{*}{\shortstack{\textcolor{cvprblue}{(1)}\\KBNet}}
    & ANCL~\cite{kim2023achieving}
    & 9.73 & 10.75 & 5.58 & 16.38
    & 56.89 & 120.30 & 13.77 & 31.85
    & 47.32 & 103.42 & 13.88 & 32.76 
    \\
    & CMP~\cite{kang2024continual}
    & 5.39 & 5.11 & 8.25 & 7.90
    & 55.92 & 117.83 & 13.74 & 31.43
    & 46.03 & 102.36 & 13.55 & 32.03
    \\
    & \textit{Ours}
    & \textbf{3.20} & \textbf{1.30} & \textbf{4.91} & \textbf{2.94}
    & \textbf{54.25} & \textbf{115.55} & \textbf{13.20} & \textbf{30.50}
    & \textbf{45.26} & \textbf{101.10} & \textbf{13.28} & \textbf{31.72}
    \\
    \midrule
    \multirow{3}*{\shortstack{\textcolor{cvprblue}{(2)}\\Uformer}}
    & Finetuned
    & 87.94 & 73.61 & 110.98 & 852.79 & 183.24 & 302.99 & 51.07 & 297.92 & 137.20 & 238.95 & 49.54 & 142.33 
    \\
    & L2P~\cite{wang2022learning}
    & 57.07 & 43.84 & 50.82 & 58.24 & 171.74 & 273.75 & 46.90 & 121.30 & 139.08 & 231.88 & 51.98 & 156.41
    \\
    & \textit{Ours}
    & \textbf{37.15} & \textbf{25.50} & \textbf{31.86} & \textbf{17.04} & \textbf{161.62} & \textbf{255.54} & \textbf{42.38} & \textbf{79.34} & \textbf{133.36} & \textbf{220.68} & \textbf{44.74} & \textbf{84.31}
    \\
    \midrule
    \multirow{3}{*}{\shortstack{\textcolor{cvprblue}{(3)}\\KBNet}}
    & ANCL~\cite{kim2023achieving}
    & 20.49 & 8.94 & 23.11 & 27.73
    & 438.05 & 1795.76 & 1.21 & 3.56
    & 503.53 & 2203.44 & 1.18 & 3.53
    \\
    & CMP~\cite{kang2024continual}
    & 15.95 & 15.47 & 6.90 & 7.39
    & 447.09 & 1887.14 & 1.09 & 3.19
    & 507.90 & 2262.46 & 1.06 & 3.21
    \\
    & \textit{Ours}
    & \textbf{4.51} & \textbf{3.10} & \textbf{2.96} & \textbf{1.88}
    & \textbf{409.90} & \textbf{1730.72} & \textbf{1.04} & \textbf{3.04}
    & \textbf{478.79} & \textbf{2138.35} & \textbf{1.01} & \textbf{3.07}
    \\
    \midrule
    \multirow{3}{*}{\shortstack{\textcolor{cvprblue}{(4)}\\KBNet}}
    & ANCL~\cite{kim2023achieving}
    & 35.10 & 35.31 & 18.13 & 10.04
    & 313.71 & 1067.35 & 18.89 & 30.39
    & 343.06 & 1129.85 & 18.66 & 30.20
    \\ 
    & CMP~\cite{kang2024continual}
    & 31.60	& 36.04 & 12.63 & 9.90
    & 307.87 & 1117.91 & 16.71 & 30.41
    & 336.08 & 1142.94 & 16.66 & 30.23 
    \\
    & \textit{Ours}
    & \textbf{20.61} & \textbf{18.75} & \textbf{9.79} & \textbf{6.25} 
    & \textbf{277.04} & \textbf{985.58} & \textbf{15.07} & \textbf{28.42} 
    & \textbf{309.57} & \textbf{1035.55} & \textbf{15.05} & \textbf{28.24}
    \\
    \midrule
    \textcolor{cvprblue}{(5)}
    & L2P~\cite{wang2022learning}
    & 69.28 & 23.25 & 81.95 & 48.78
    & 519.72 & 1458.78 & 25.65 & 36.21
    & 470.84 & 1407.23 & 25.38 & 35.45
    \\
    Uformer
    & \textit{Ours}
    & \textbf{45.42} & \textbf{7.67} & \textbf{46.18} & \textbf{22.05}
    & \textbf{451.08} & \textbf{1252.88} & \textbf{22.34} & \textbf{32.00}
    & \textbf{401.95} & \textbf{1220.67} & \textbf{21.97} & \textbf{31.63}
    \\
    \midrule
\end{tabular}
}
\vspace{-3mm}
\caption{\textbf{Additional quantitative results} comparing to recent baselines on indoor, outdoor, and mixed sequences with backbone as denoted:\\
\textcolor{cvprblue}{(1,2)} \underline{Indoor}: NYUv2 $\rightarrow$ ScanNet $\rightarrow$ VOID \,\,\,\,\, \textcolor{cvprblue}{(3)} \underline{Outdoor}: KITTI $\rightarrow$ Waymo $\rightarrow$ VKITTI \,\,\,\,\, \textcolor{cvprblue}{(4,5)} \underline{Mixed}: KITTI $\rightarrow$ NYUv2 $\rightarrow$ Waymo
}
\vspace{-1mm}
\label{tab:additional_quantitative}
\end{table*}

\section{Transformer Experiments}

In the main paper, we stated that ProtoDepth is applicable to any model with a latent space, including both CNNs and transformers.
To explore the applicability of ProtoDepth to transformer-based architectures, we adapted Uformer~\cite{wang2022uformer}, a simple encoder-decoder model consisting entirely of transformer blocks, for depth completion. The model takes as input patchified versions of the image and sparse depth, where inputs from each modality are split into $14\times14$ patches and embedded as $N\times C$ tokens. We adapted Uformer for depth completion by implementing a dual-encoder structure, with one encoder processing image tokens and the other processing sparse depth tokens. Each encoder contains four transformer blocks. After being processed by the encoders, the tokens from both modalities are concatenated and fed into a shared decoder with four additional transformer blocks. Consistent with the CNN-based models used in the main paper, skip connections are included between each encoder block and its corresponding decoder block, allowing multi-scale features to flow between the encoders and decoder.

For ProtoDepth-A and ProtoDepth, we implemented our method in the exact same way as we do for CNN-based models, applying prototype sets to the latent space layers, i.e., the bottleneck and skip connections. The prototype sets learn global (multiplicative) and local (additive) biases for each layer, adapting the frozen transformer layers to each new dataset while mitigating forgetting. This demonstrates that ProtoDepth is fully architecture-agnostic and can be seamlessly applied to both CNNs and transformers.

A notable inclusion in this section is the prompt-based method L2P~\cite{wang2022learning} (Learning to Prompt), which serves as a representative baseline for prompt-based methods. Prompt-based continual learning methods were not included in the main experiments because all existing unsupervised depth completion models are CNN-based, and prompt-based approaches, which operate by prepending prompts to tokenized inputs, are not applicable to CNNs, which operate directly on images without tokenization, which prevents the straightforward insertion of prompts into the input space. However, with the implementation of Uformer, a transformer-based model, we are now able to evaluate L2P, which is a foundational method for prompt-based continual learning.

For L2P, we implement the method as described in the original paper. Specifically, we use a prompt pool of size $M=20$ and select $N=5$ prompts for each input during training and inference. To adapt L2P for depth completion, we implement their loss term, which pulls selected keys closer to their corresponding queries, and incorporate it into our overall loss function (Eq. (1) in the main paper) with a weight of 0.5, as suggested in~\cite{wang2022learning}. To evaluate in the domain-agnostic setting, where dataset identity is withheld at test time, we train $M=20$ new prompts for each new dataset during continual training. At test-time, the model queries all existing learned prompts.

\section{Additional Experiments}

In \cref{tab:additional_quantitative}-\textcolor{cvprblue}{(2)}, we compare to L2P [Wang et al., CVPR '22]~\cite{wang2022learning}, a prompt-based method, where we adapt Uformer for unsupervised depth completion as no transformer-based model currently exists for this task. We have added comparisons to ANCL [Kim et al., CVPR '23]~\cite{kim2023achieving}, an architecture-based method, and CMP [Kang et al., CVPR '24]~\cite{kang2024continual}, a rehearsal-based method, on the indoor \cref{tab:additional_quantitative}-\textcolor{cvprblue}{(1)} and outdoor \cref{tab:additional_quantitative}-\textcolor{cvprblue}{(3)} sequences using the KBNet backbone. ProtoDepth-A (\textit{Ours}) outperforms all of these recent methods, reaffirming our findings.

In \cref{tab:additional_quantitative}-\textcolor{cvprblue}{(4,5)}, we add experiments in a mixed setting, where the dataset sequence transitions from outdoor to indoor and back to outdoor. We compare to ANCL, CMP, and L2P in this mixed setting and show that ProtoDepth-A outperforms all of these recent methods.

\cref{tab:foundation} shows that recent depth estimation unified/foundation models, Depth Pro [Bochkovskii et al., 2024]~\cite{bochkovskii2024depth} and Depth Anything [Yang et al., CVPR '24]~\cite{yang2024depth} (fit to metric scale via median scaling) do \textit{not} outperform ProtoDepth-A (NYU $\rightarrow$ VOID) when evaluated on \textbf{VOID}. This validates the advantage of our method over direct depth estimation. 
Also of note, Depth Pro and Depth Anything are supervised and semi-supervised, while we are unsupervised.

In continual learning, joint training a larger model (e.g., transformer) on all datasets simultaneously serves as a performance upper bound. \cref{tab:upper_bound} shows that ProtoDepth-A achieves comparable mean performance to this upper bound on \textbf{\{KITTI, Waymo, VKITTI\}} using the adapted Uformer. Importantly, we address the scientific question of learning in a sequential manner, where one does not have access to all data at once or must learn a new dataset without breaking backwards-compatibility -- a common real-world scenario.

Improved generalization to unseen datasets in the intersection of observed domains helps to motivate our method. \cref{tab:generalization} shows generalization to \textbf{nuScenes} (outdoor) after training on KITTI $\rightarrow$ Waymo $\rightarrow$ VKITTI. ProtoDepth-A outperforms joint training, ANCL, and CMP, demonstrating its ability to leverage domain-specific prototypes to enhance zero-shot generalization.

\begin{table}[!t]
\normalsize
\centering
\setlength\tabcolsep{4pt}
\resizebox{\columnwidth}{!}{
\begin{tabular}{l@{\hspace{10pt}} c c c c}
    \toprule
    {} & MAE & RMSE & iMAE & iRMSE \\ [-0.5mm]
    \midrule
    Depth Anything~\cite{yang2024depth} & 49.22 & 88.74 & 21.22 & 51.22 \\
    Depth Pro~\cite{bochkovskii2024depth} & 43.06 & 93.36 & 20.80 & 52.24 \\
    \textit{Ours} & \textbf{33.66} & \textbf{86.99}	& \textbf{17.48} & \textbf{43.02} \\
    \midrule
\end{tabular}
}
\vspace{-3mm}
\caption{Comparison against depth estimation foundation models.
}
\vspace{-1mm}
\label{tab:foundation}
\end{table}

\begin{table}[!t]
\normalsize
\centering
\setlength\tabcolsep{4pt}
\resizebox{\columnwidth}{!}{
\begin{tabular}{l@{\hspace{30pt}} c c c c}
    \toprule
    {} & MAE & RMSE & iMAE & iRMSE \\ [-0.5mm]
    \midrule
    \textit{Ours} & 686.86 & \textbf{2024.42} & 1.58 & \textbf{3.52} \\     
    Upper Bound & \textbf{671.95} & 2231.97 & \textbf{1.34} & \textbf{3.52} \\
    \midrule
\end{tabular}
}
\vspace{-3mm}
\caption{Comparison against joint training (upper bound).
}
\vspace{-1mm}
\label{tab:upper_bound}
\end{table}

\begin{table}[!t]
\normalsize
\centering
\setlength\tabcolsep{4pt}
\resizebox{\columnwidth}{!}{
\begin{tabular}{l@{\hspace{30pt}} c c c c}
    \toprule
    {} & MAE & RMSE & iMAE & iRMSE \\ [-0.5mm]
    \midrule
    Joint Training & 2800.27 & 6284.63 & 6.06 & 11.23 \\
    ANCL~\cite{kim2023achieving} & 2753.07 & 6195.09 & 5.69 & 10.86 \\
    CMP~\cite{kang2024continual} & 2885.82 & 6234.33 & 7.12 & 13.57 \\
    \textit{Ours} & \textbf{2697.47} & \textbf{5966.57} & \textbf{5.40} & \textbf{10.58} \\
    \midrule
\end{tabular}
}
\vspace{-3mm}
\caption{Zero-shot generalization to nuScenes.
}
\vspace{-3mm}
\label{tab:generalization}
\end{table}

\section{Dataset Details} \label{sec:datasets}

\hspace*{1pc}\textit{Indoor datasets}: The \textbf{NYU Depth V2}~\cite{silberman2012indoor} (``NYUv2'') dataset comprises 464 diverse indoor scenes from residential, office, and commercial environments captured using a Microsoft Kinect. It contains approximately 400,000 aligned RGB and depth image pairs with a resolution of $640 \times 480$. About 1,500 points are sampled for each sparse depth map using the Harris corner detector~\cite{harris1988combined}. This dataset serves as a standard benchmark for indoor depth estimation tasks. For our indoor dataset sequence, we utilize NYUv2 as the initial dataset $\mathcal{D}_1$ for pretraining our depth completion models that are subsequently applied to indoor continual learning scenarios. The \textbf{VOID}~\cite{wong2020unsupervised} dataset presents sparse depth maps with $\approx0.5\%$ density ($\approx$1,500 points), alongside RGB frames from various indoor settings such as laboratories, classrooms, and gardens, totaling approximately 58,000 frames ($640 \times 480$) captured via XIVO~\cite{fei2019geo}. VOID is designed to address challenges in areas with minimal texture and significant camera motion, key factors for assessing robustness in indoor depth completion tasks. \textbf{ScanNet} \cite{dai2017scannet}, a comprehensive indoor dataset, encompasses over 2.5 million frames paired with RGB-D data. Depth frames in ScanNet are captured at a resolution of $640 \times 480$ pixels, whereas the color frames have a higher resolution of $1296 \times 968$ pixels. Again, we use the Harris corner detector~\cite{harris1988combined} to subsample $\approx$ 1,500 points for the sparse depth maps.
We use a subset of the dataset with approximately 250,000 frames across 706 scenes.
For all indoor datasets, we use a training crop size of $416 \times 576$. 
For evaluation, depth values across all of these indoor datasets are constrained between 0.2 and 5 meters. 

\textit{Outdoor datasets:} The \textbf{KITTI}~\cite{uhrig2017sparsity} dataset is an established benchmark in autonomous driving that comprises over 93,000 stereo image pairs with a resolution of $1240 \times 376$ and sparse LiDAR depth maps ($\approx5\%$ density), all synchronized and captured across diverse urban and rural landscapes using a Velodyne LiDAR sensor. KITTI is the initial dataset $\mathcal{D}_1$ for pretraining our depth completion models for the outdoor dataset sequence. The \textbf{Waymo Open Dataset}~\cite{sun2020scalability} (``Waymo'') provides roughly 230,000 high-resolution frames ($1920 \times 1280$ and $1920 \times 1040$) along with LiDAR point clouds, captured from scenes that encompass a broad spectrum of driving scenarios and conditions. For Waymo, the depth values during evaluation are capped between 0.001 and 80 meters and during training, a crop size of $800 \times 640$ is employed. The \textbf{Virtual KITTI}~\cite{gaidon2016virtual} (``VKITTI'') dataset offers synthetic, altered re-creations of KITTI scenes captured from virtual worlds created in Unity, with over 21,000 frames at $1242 \times 375$ resolution and dense ground truth depth, facilitating the study of domain adaptation. We apply synthetic weather conditions and view rotations to simulate domain shifts that lead to forgetting. For KITTI and VKITTI, we restrict the depth values during evaluation to between 0.001 and 100 meters and utilize a depth cropping of $240 \times 1216$. During training, we use a crop size of $320 \times 768$.

Given the differences in image resolutions, crop sizes, and evaluation depths, in addition to the different types of scenes captured and sensors used to collect the datasets, we observe large domain gaps between datasets within each sequence, motivating the need for continual learning. We will release code for reproducibility.

\section{Depth Completion Metrics}

When we reference depth completion metrics in the main paper, we specifically mean the \textit{error} metrics outlined below and formulated in \cref{tab:error_metrics_depth_completion}. The metrics include Mean Absolute Error (MAE), Root Mean Squared Error (RMSE), Inverse Mean Absolute Error (iMAE), and Inverse Root Mean Squared Error (iRMSE). MAE measures the average $L1$ difference between predicted and ground-truth depths, providing a straightforward indication of prediction accuracy. RMSE measures $L2$ difference which gives higher weight to larger errors, making it sensitive to outliers and thus a robust measure for practical applications. iMAE and iRMSE, on the other hand, are particularly useful for scenarios where errors in smaller depth values are more critical, as they focus on the relative error in inverse depth. Collectively, these metrics allow for a comprehensive evaluation of a model's capability to predict depth from input data under varied environmental settings, e.g., indoor and outdoor. We note that lower values indicate better performance for all four error metrics. All results are reported in `mm' (millimeters) unless otherwise specified, providing a clear metric standardization.

\begin{table}[b]
\centering
    \scalebox{1.0}{
    \begin{tabular}{l l}
        \midrule
            Metric & Definition \\ \midrule
            MAE\,$\downarrow$ &$\frac{1}{|\Omega|} \sum_{x\in\Omega} |\hat d(x) - d(x)|$ \\
            RMSE\,$\downarrow$ & $\big(\frac{1}{|\Omega|}\sum_{x\in\Omega}|\hat d(x) - d(x)|^2 \big)^{1/2}$ \\
            iMAE\,$\downarrow$ & $\frac{1}{|\Omega|} \sum_{x\in\Omega} |1/ \hat d(x) - 1/d(x)|$ \\
            iRMSE\,$\downarrow$ & $\big(\frac{1}{|\Omega|}\sum_{x\in\Omega}|1 / \hat d(x) - 1/d(x)|^2\big)^{1/2}$ \\ 
            \midrule
        \end{tabular}
    }
    \caption{
        \textbf{Error metrics for depth completion.} These metrics evaluate the accuracy of predicted depth values $\hat{d}(x)$ compared to ground truth depth values $d(x)$ over the set of pixels $\Omega$.}
\label{tab:error_metrics_depth_completion}
\end{table}

\begin{table*}[t]
\scriptsize
\centering
\setlength\tabcolsep{2pt}
\resizebox{1.0\textwidth}{!}{%
\begin{tabular}{c c c c@{\hspace{8pt}} c c c c@{\hspace{8pt}} c c c c}
    \toprule
    & & & & \multicolumn{4}{c}{Waymo} & \multicolumn{4}{c}{VKITTI} \\
    \midrule
    Method & $N^{(I)}$ & $N^{(z)}$ & \# Params & MAE & RMSE & iMAE & iRMSE  & MAE & RMSE & iMAE & iRMSE \\ 
    \midrule
    Pretrained & - & - & 0M (0\%)
    & 3930.68 & 6405.75 & 9.55 & 14.34
    & 10527.70 & 18086.22 & 17.45 & 31.50
    \\
    \midrule
    \multirow{7.75}*{\methodabbr} & 
    1 & 1 & 0.24M (3.5\%)
    & 587.92 & 1900.96 & 1.41 & 2.96
    & 937.18 & 4027.53 & 1.92 & 5.82
    \vspace{-0.8ex}
    \\
    & & & 
    & \tiny{$\pm$61.20} & \tiny{$\pm$145.34} & \tiny{$\pm$0.12} & \tiny{$\pm$0.17}
    & \tiny{$\pm$60.31} & \tiny{$\pm$47.08} & \tiny{$\pm$0.38} & \tiny{$\pm$0.42}
    \\
    & 10 & 10 & 0.25M (3.7\%)
    & 524.76 & 1667.74 & 1.28 & 2.74
    & 686.22 & 3638.20 & 0.90 & 3.50
    \vspace{-0.8ex}
    \\
    & & & 
    & \tiny{$\pm$37.18} & \tiny{$\pm$27.98} & \tiny{$\pm$0.06} & \tiny{$\pm$0.03}
    & \tiny{$\pm$3.42} & \tiny{$\pm$12.29} & \tiny{$\pm$0.04} & \tiny{$\pm$0.07}
    \\
    & 25 & 10 & 0.27M (3.9\%)
    & \textbf{483.92} & \textbf{1656.33} & \textbf{1.19} & \textbf{2.68 }
    & \textbf{676.28} & \textbf{3608.42} & \textbf{0.80} & \textbf{3.25}
    \vspace{-0.8ex}
    \\
    & & & 
    & \tiny{$\pm$27.59} & \tiny{$\pm$16.34} & \tiny{$\pm$0.04} & \tiny{$\pm$0.02}
    & \tiny{$\pm$4.64} & \tiny{$\pm$16.61} & \tiny{$\pm$0.07} & \tiny{$\pm$0.24}
    \\
    & 25 & 25 & 0.28M (4.0\%)
    & 508.60 & 1688.09 & 1.23 & 2.72 
    & 680.65 & 3614.61 & 0.87 & 3.51
    \vspace{-0.8ex}
    \\
    & & & 
    & \tiny{$\pm$20.36} & \tiny{$\pm$10.88} & \tiny{$\pm$0.04} & \tiny{$\pm$0.03}
    & \tiny{$\pm$3.40} & \tiny{$\pm$14.82} & \tiny{$\pm$0.05} & \tiny{$\pm$0.19}
    \\
    & 100 & 100 & 0.38M (5.5\%)
    & 522.39 & 1711.44 & 1.27 & 2.76
    & 686.89 & 3635.01 & 0.93 & 3.53
    \vspace{-0.8ex}
    \\
    & & & 
    & \tiny{$\pm$50.06} & \tiny{$\pm$72.41} & \tiny{$\pm$0.10} & \tiny{$\pm$0.09}
    & \tiny{$\pm$5.45} & \tiny{$\pm$27.57} & \tiny{$\pm$0.09} & \tiny{$\pm$0.08}
    \\
    \midrule
\end{tabular}
}
\vspace{-3mm}
\caption{\textbf{Sensitivity study} of prototype set sizes ($N^{(I)}$ and $N^{(z)}$) on \methodabbr using KBNet for outdoor datasets (Waymo and VKITTI). KBNet is pretrained on the initial dataset (KITTI). Parameter overhead is reported as a percentage of the full KBNet model's parameters. Smaller set sizes show suboptimal performance due to insufficient capacity to capture feature diversity, while larger set sizes also degrade performance, likely from overfitting and learning noise.
}
\vspace{-3mm}
\label{tab:outdoor_set_sizes}
\end{table*}

The results of our experiments are shown in \cref{tab:additional_quantitative}, which compares ProtoDepth, ProtoDepth-A (agnostic setting), L2P, and full finetuning (``Finetuned'') on the indoor dataset sequence. ProtoDepth achieves superior performance across all metrics, with zero forgetting in the incremental setting, with one exception: ProtoDepth-A outperforms ProtoDepth in one measure, SPTO for iRMSE, highlighting the benefits of its generalization capability. This result is consistent with our earlier observations: by allowing the model to select domain descriptors and prototype sets dynamically at test time, ProtoDepth-A can leverage features from overlapping domains to improve performance on ambiguous samples. This flexibility enables better generalization, which, in certain scenarios, can lead to improved outcomes compared to the fixed domain identity approach used in ProtoDepth.

Notably, ProtoDepth-A outperforms L2P in the agnostic setting, demonstrating the strength of prototype-based adaptation compared to prompt-based approaches. While L2P shows improvements over finetuning, it performs less well than ProtoDepth, which can be attributed to a fundamental limitation of prompt-based methods. These methods rely on learnable prompts or tokens to adapt frozen vision transformer models for continual learning, but there is no natural scale at which to discretize images or choose an appropriate prompt size, unlike the discrete text tokens used in natural language processing. In contrast, ProtoDepth's prototype-based approach eliminates the need for tokenized inputs, enabling it to operate directly in the latent feature space. This flexibility not only enhances its adaptability across diverse datasets but also allows it to be applied seamlessly to both transformers and convolutional neural networks, which are prevalent in unsupervised depth completion.

\section{Outdoor Prototype Set Sizes}

We extend our investigation of prototype set sizes (i.e., number of prototypes) for the image and sparse depth layers (denoted as $N^{(I)}$ and $N^{(z)}$, respectively) to the outdoor dataset sequence. The results of these experiments are presented in \cref{tab:outdoor_set_sizes}. Based on the findings, we select $N^{(I)} = 25$ and $N^{(z)} = 10$ for the main experiments on the outdoor dataset sequence. Smaller set sizes demonstrate suboptimal performance, as they lack the capacity to adequately capture the diversity of features across datasets. Larger set sizes also result in performance degradation, likely due to the additional parameters learning noise and overfitting to the training data. The best performance is achieved when $N^{(I)} > N^{(z)}$, aligning with our observations in the indoor experiments. This can be attributed to the larger distributional shift between scenes in the image modality compared to the sparse depth modality~\cite{park2024testtime}. For the bottleneck layer, which fuses features from both modalities, we again use $N^{(I)}$ as the prototype set size. As a baseline, we also report the performance of the frozen base model pretrained on KITTI (``Pretrained''), which has no additional parameters or further training. The poor results highlight the necessity of continual learning to adapt to non-stationary data distributions. 
For both indoor and outdoor settings, the prototype set size analysis is conducted using the KBNet model; we adopt the same prototype set sizes for all other models, as they all have a similar number of parameters.

\section{Additional Qualitative Analysis}

\begin{figure*}[t]
  \centering
  \includegraphics[width=1.0\linewidth]{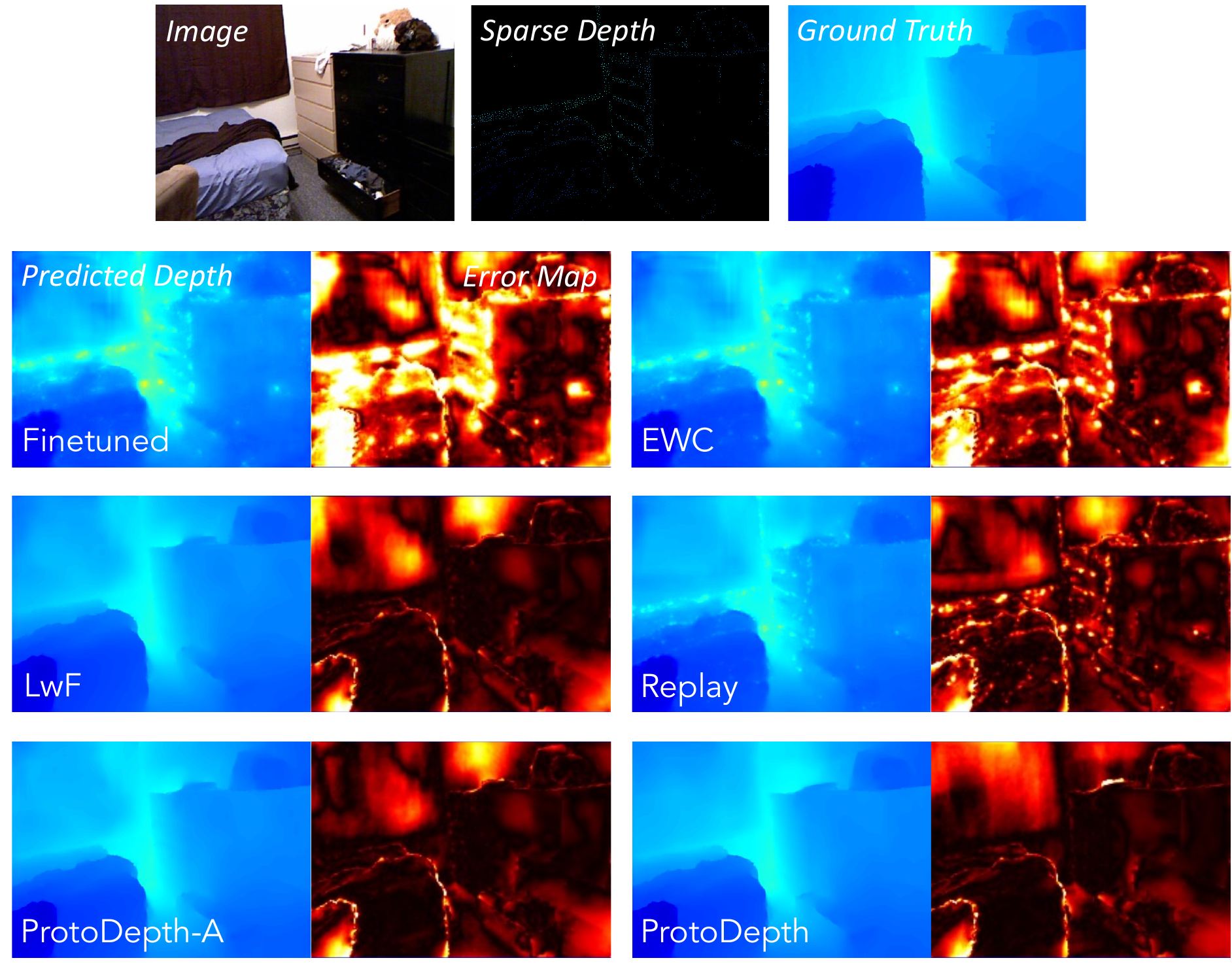}
  \vspace{-6mm}
  \caption{\textbf{Qualitative comparison} (1 of 2) of \methodabbr and baseline methods using FusionNet on \textbf{NYUv2} after continual training on \textbf{ScanNet}. \textit{Top row:} Input sample from NYUv2. \textit{Following rows:} Output depth and error maps (relative to ground-truth) of same sample from NYUv2 after continual training on ScanNet using each continual learning method.}
  \label{fig:qualitative_indoor1}
  \vspace{-3mm}
\end{figure*}

\begin{figure*}[t]
  \centering
  \includegraphics[width=1.0\linewidth]{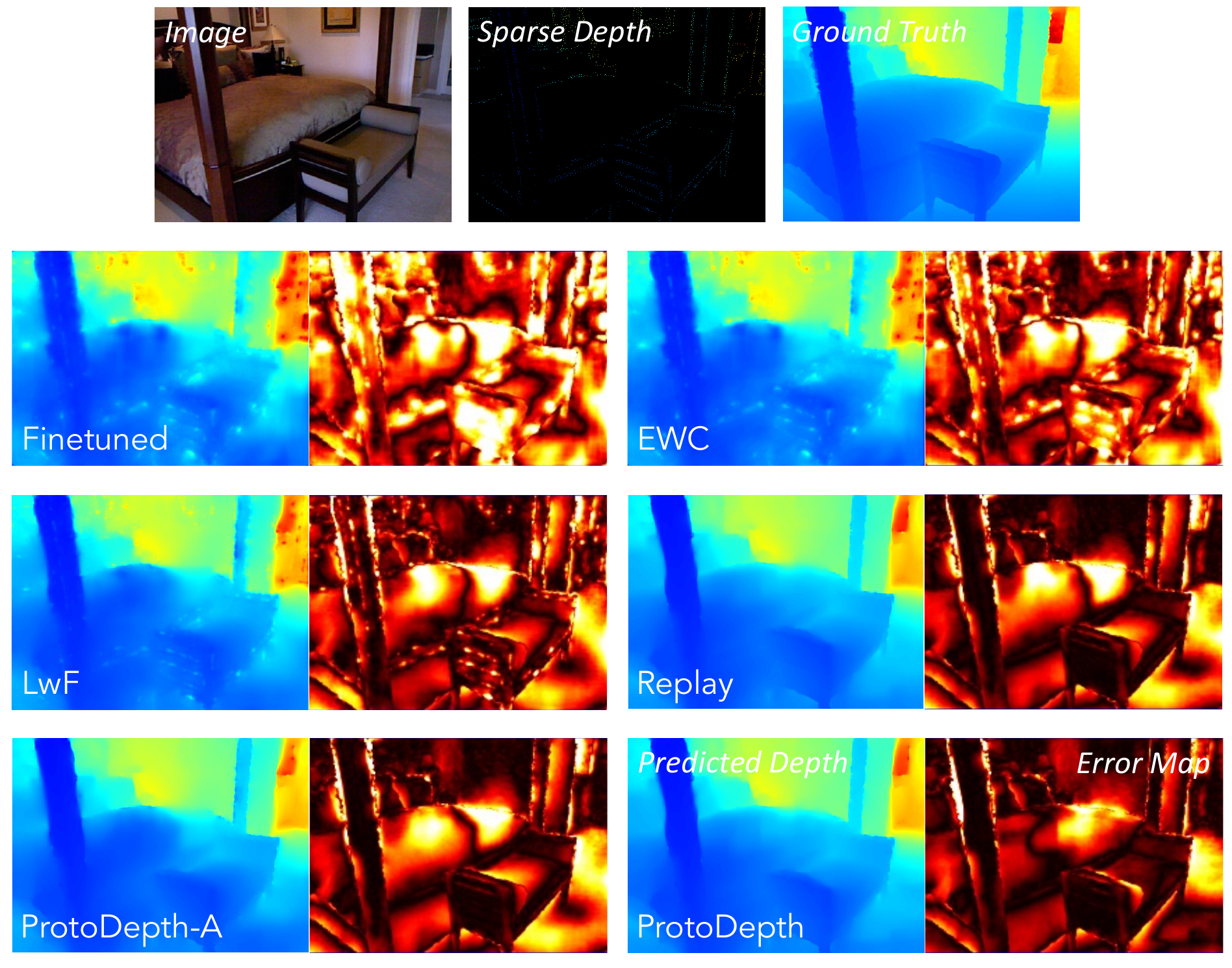}
  \vspace{-6mm}
  \caption{\textbf{Qualitative comparison} (2 of 2) of \methodabbr and baseline methods using FusionNet on \textbf{NYUv2} after continual training on \textbf{ScanNet}. \textit{Top row:} Input sample from NYUv2. \textit{Following rows:} Output depth and error maps (relative to ground-truth) of same sample from NYUv2 after continual training on ScanNet using each continual learning method.}
  \label{fig:qualitative_indoor2}
  \vspace{-3mm}
\end{figure*}

To illustrate the reduced forgetting achieved by ProtoDepth, we provide a qualitative comparison of depth predictions and error maps for all baseline methods on input samples from NYUv2 after continual training on ScanNet (\cref{fig:qualitative_indoor1} and \cref{fig:qualitative_indoor2}). These figures demonstrate how ProtoDepth and ProtoDepth-A consistently outperform the baselines, specifically in reconstructing crowded indoor scenes with sparse depth measurements and challenging lighting conditions.

In \cref{fig:qualitative_indoor1}, baseline methods such as Finetuned and EWC exhibit substantial forgetting, resulting in high error concentrations. Finetuned, in particular, struggles to retain photometric priors learned from NYUv2, evident in the poor reconstruction of furniture edges and flat areas with depth gradients. Replay performs marginally better but still fails to recover fine details, as its rehearsal mechanisms are insufficient to address the large distributional shift between NYUv2 and ScanNet. LwF shows improved performance, with fewer errors compared to Finetuned, EWC, and Replay. However, it fails to accurately reconstruct regions with sparse depth measurements (see Sparse Depth), such as the curtain.

ProtoDepth and ProtoDepth-A, on the other hand, produce high-fidelity depth predictions. ProtoDepth benefits from its prototype-based adaptation, effectively preserving features from NYUv2 while adapting to ScanNet. Notably, ProtoDepth-A exhibits comparable performance and even outperforms ProtoDepth in reconstructing certain regions, such as the smooth surface of the curtain. This improvement is due to ProtoDepth-A’s generalization capability, which allows it to dynamically select prototype sets from overlapping domains based on the affinity of domain descriptors, thereby enhancing its ability to handle ambiguous inputs.

\cref{fig:qualitative_indoor2} reinforces these observations with a second example. Once again, baseline methods exhibit significant forgetting, with Finetuned, EWC, and LwF producing poor depth predictions. In contrast, ProtoDepth and ProtoDepth-A produce high-fidelity reconstructions. The well-defined edges between the furniture, floor, and walls in their predictions highlight their ability to preserve learned features while adapting to new domains. ProtoDepth-A, in particular, demonstrates its generalization strength by leveraging overlapping domain features to improve predictions in certain areas, such as the bedpost edges.

Overall, these qualitative results underscore the ability of ProtoDepth to mitigate catastrophic forgetting and produce high-fidelity depth predictions. By effectively combining domain-specific adaptation and cross-domain generalization, ProtoDepth-A outperforms baseline methods, even under significant domain shifts between NYUv2 and ScanNet.

\section{Training Time Comparison}

\cref{tab:training_time} presents the training time per epoch for each continual learning method on both indoor (ScanNet and VOID) and outdoor (Waymo and VKITTI) datasets using KBNet. These experiments were conducted with a fixed batch size of 12 for indoor datasets and 8 for outdoor datasets, on a single NVIDIA GeForce RTX 3090 GPU. This standardized setup ensures a fair comparison across all methods. The training times vary across datasets because they are measured per epoch, and each training set contains a different number of frames, as detailed in \cref{sec:datasets}.

\begin{table}[b]
\normalsize
\centering
\setlength\tabcolsep{4pt}
\begin{tabular}{l c c c c}
    \toprule
    & \multicolumn{4}{c}{Training Time per Epoch (mins)} \\
    \midrule
    Method & ScanNet & VOID & Waymo & VKITTI \\
    \midrule 
    Finetuned & 165.8 & 35.4 & 84.7 & 17.3 \\
    EWC & 168.2 & 35.9 & 85.0 & 18.5 \\
    LwF & 170.7 & 38.1 & 85.4 & 20.3 \\
    Replay & 182.9 & 40.4 & 88.8 & 23.0 \\
    \textit{\methodabbr-A} & 92.5 & 17.9 & 40.3 & 10.7 \\
    \textit{\methodabbr} & 85.3 & 15.7 & 37.9 & 9.6 \\
    \midrule
\end{tabular}
\vspace{-3mm}
\caption{\textbf{Training times} (minutes per epoch) with KBNet for each continual learning method on both indoor and outdoor datasets.
}
\vspace{-3mm}
\label{tab:training_time}
\end{table}

ProtoDepth and ProtoDepth-A demonstrate significant improvements in computational efficiency, with training times roughly half those of the baseline methods. This efficiency can be attributed to ProtoDepth’s approach of freezing the backbone model and training only the prototype sets, which are applied to the latent space layers (i.e., bottleneck and skip connections). Thus, backpropagation computations are restricted to parameters from the output layer back only to the latent space layers. Since the parameters involved are approximately half of the total parameters, ProtoDepth requires fewer gradient computations compared to methods like EWC, LwF, and Replay that calculate gradients and update parameters across the entire model.

ProtoDepth achieves slightly faster training times than ProtoDepth-A. This difference arises because ProtoDepth-A requires additional computations to train the domain descriptors, which involves calculating and optimizing cosine similarity between sample descriptors and domain descriptors during training. ProtoDepth avoids this step, resulting in a small yet consistent reduction in training time.

Among the baseline methods, Finetuned is the fastest, training slightly faster than EWC, LwF, and Replay. This is because finetuning does not involve the additional regularization or distillation used by EWC and LwF, nor does it use a memory buffer like Replay. However, the simplicity of full finetuning comes at the cost of increased catastrophic forgetting, as evidenced by its consistently poor performance in the main experiments.

The reduced training times of ProtoDepth and ProtoDepth-A are particularly important for real-world applications, where computational efficiency is crucial. By restricting updates to the latent space, ProtoDepth not only reduces computational overhead but also does so while achieving state-of-the-art performance. This efficiency is critical for resource-constrained environments, or scenarios requiring fast adaptation to new datasets. These results highlight ProtoDepth’s ability to deliver both high performance and practical advantages in training time, underscoring its suitability for continual learning tasks.

\section{More Ablation Studies}

To further evaluate the importance of prototype sets in ProtoDepth, we conduct additional ablation studies to assess the impact of removing prototype sets from different modalities and latent space layers. Specifically, we analyze the role of prototype sets applied to the image features, sparse depth features, and the bottleneck features. The results, shown in \cref{tab:additional_ablations}, are evaluated on ScanNet (indoor dataset) and Waymo (outdoor dataset) using KBNet.

The results highlight that removing prototype sets from any of these components significantly degrades performance. When image prototype sets are ablated, we observe a sharp increase in both MAE and RMSE, particularly for ScanNet, where MAE rises from 14.59 to 35.06. This degradation demonstrates the importance of capturing domain-specific biases in image features, as images undergo larger distributional shifts between domains compared to sparse depth, such as changes in lighting, textures, and color distributions.

Similarly, removing the sparse depth prototype sets also results in noticeable performance drops, with MAE increasing from 14.59 to 32.07 for ScanNet. While sparse depth features may exhibit smaller distributional shifts compared to image features, these features are crucial for anchoring the model to the metric scale of the depth predictions. Without the sparse depth prototypes, the model struggles to adapt effectively to the unique distribution of sparse point clouds in each new dataset.

The bottleneck prototype sets play a critical role as well, as they adapt the fused representations of both image and sparse depth modalities. Ablating the bottleneck prototypes leads to performance degradation, although the impact is less severe than removing the image or sparse depth prototypes. For instance, MAE increases from 14.59 to 19.03 for ScanNet when bottleneck prototypes are removed. This suggests that while the bottleneck prototypes contribute to the overall performance, much of the adaptation occurs in the modality-specific layers.

Notably, when all prototype sets are included (no ablations), ProtoDepth achieves the best performance across both datasets, with significantly lower error metrics compared to any ablated configuration. These results validate the design choice of applying prototype sets to both modality-specific features (image and sparse depth) and their fused representations (bottleneck).

\begin{table}[t]
\scriptsize
\centering
\setlength\tabcolsep{2pt}
\resizebox{\columnwidth}{!}
{%
\begin{tabular}{l@{\hspace{8pt}} c c@{\hspace{8pt}} c c}
    \toprule
    & \multicolumn{2}{c}{ScanNet\hspace*{7pt}} & \multicolumn{2}{c}{Waymo} \\
    \midrule
    Ablated Component & MAE & RMSE & MAE & RMSE\\
    \midrule 
    image prototype sets & 35.06 & 88.23 & 542.16 & 1703.01 \\
    sparse depth prototype sets & 32.07 & 84.39 & 537.37 & 1762.31 \\
    bottleneck prototype sets & 19.03 & 60.32 & 502.21 & 1680.87 \\
    \midrule
    no ablations & 14.59 & 42.20 & 486.95 & 1664.18 \\
    \midrule
\end{tabular}
}
\vspace{-3mm}
\caption{\textbf{Ablation studies} on prototype sets for different modalities using KBNet for indoor (ScanNet) and outdoor (Waymo).
}
\vspace{-3mm}
\label{tab:additional_ablations}
\end{table}

\section{Discussion}

Accurate 3D reconstruction~\cite{lao2024sub, upadhyay2023enhancing, xie2023sparsefusion} is crucially important for applications that rely on precise perception of surrounding environments~\cite{zeng2024rsa, xia2023quadric}.
One key challenge in this domain is monocular depth estimation (MDE)~\cite{bochkovskii2024depth, lao2024depth, wong2019bilateral, wong2020targeted, wu2024augundo, yang2024depth}, which aims to recover metric depth from a single image. However, MDE is fundamentally challenging due to scale ambiguity, making it an inherently ill-posed problem.
To overcome this challenge, synchronized complementary modalities---such as LiDAR~\cite{ezhov2024all, wong2020unsupervised, wong2021unsupervised}, radar~\cite{singh2023depth}, inertial sensors~\cite{fei2019geo}, additional cameras~\cite{wong2021stereopagnosia,berger2022stereoscopic}, and even language~\cite{zeng2024wordepth, zeng2024priordiffusion}---can provide additional cues to resolve scale ambiguity.
In particular, LiDAR offers high-precision depth measurements that are relatively dense compared to other time-of-flight sensors such as radar, making it a valuable modality for resolving scale ambiguity and enhancing metric depth estimation accuracy.
This task of LiDAR-Camera depth estimation, specifically, is commonly referred to as depth completion~\cite{wong2021adaptive, wong2021learning, yang2019dense, liu2022monitored}.
In our work, ProtoDepth, we introduce an unsupervised continual depth completion~\cite{gangopadhyay2024uncle} framework that leverages prototypes to continuously learn in challenging and dynamic environments.
Unlike traditional approaches that rely on fully supervised training on stationary datasets, ProtoDepth adapts continuously across domains, demonstrating improved generalization without the need for expensive, inaccurate ground truth.
Our comprehensive results demonstrate that ProtoDepth effectively mitigates catastrophic forgetting for depth completion, making it a promising solution for real-world applications in autonomous driving, augmented/virtual reality, robotics, and general scene understanding.

\end{document}

%% file: preamble.tex
\usepackage[dvipsnames]{xcolor}

\usepackage{url}

\usepackage{algorithm}
\usepackage{algorithmic}

\usepackage[utf8]{inputenc} 
\usepackage[T1]{fontenc}    
\usepackage{multirow}
\usepackage{tikz}

\usepackage{booktabs}
\usepackage[font=small]{caption}

\usepackage{url}            
\usepackage{booktabs}       
\usepackage{amsfonts}       
\usepackage{nicefrac}       
\usepackage{microtype}      
\usepackage{xcolor}         
\usepackage{amsmath}
\usepackage{pifont}
\usepackage{adjustbox}
\usepackage{graphicx}

\definecolor{PaleBlue}{RGB}{100, 160, 190}
\definecolor{Purple}{RGB}{153, 51, 255}